%% file: cvpr.tex
\pdfoutput=1
\documentclass[final]{cvpr}

\usepackage{times}
\usepackage{epsfig}
\usepackage{graphicx}
\usepackage[english]{babel}
\usepackage[utf8]{inputenc}
\usepackage{algorithm}
\usepackage[noend]{algpseudocode}
\usepackage{amsmath,amssymb} % define this before the line numbering.
\usepackage{graphics}
\usepackage{algorithm} 
\usepackage{algpseudocode}
\usepackage{amssymb}% http://ctan.org/pkg/amssymb
\usepackage{pifont}% http://ctan.org/pkg/pifont
\usepackage{subcaption}
\usepackage{stmaryrd}
\usepackage{booktabs}
\usepackage{microtype}
\usepackage[dvipsnames]{xcolor}

\usepackage{hyperref}
\hypersetup{breaklinks=true,colorlinks}

\DeclareMathOperator*{\argmin}{arg\,min}

\def\cvprPaperID{10362} % *** Enter the CVPR Paper ID here
\def\confYear{CVPR 2021}

\title{Divide-and-Conquer for Lane-Aware Diverse Trajectory Prediction}

\author{Sriram Narayanan$^1$ \and 
Ramin Moslemi$^1$ \and
Francesco Pittaluga$^1$ \and
Buyu Liu$^1$ \and
Manmohan Chandraker$^{1,2}$ \\[3mm]
$^1$NEC Labs America, $^2$UC San Diego
}
\begin{document}

%%%%%%%%% TITLE
% \title{SMART++: Hybrid Trajectory Prediction with Lane Anchors}

% \author{Sriram N N, 
% Ramin Moslemi, 
% Francesco Pittaluga, 
% Buyu Liu and 
% Manmohan Chandraker \\
% NEC Labs America \\
% % {\tt\small \{sriram, rmoslemi, francescopittaluga, buyu, manu\}@nec-labs.com}
% }

% \and

% \author{Sriram N N\\
% NEC Labs America\\
% San Jose, USA\\ 
% {\tt\small sriram@nec-labs.com} \and
% Ramin Moslemi \\
% NEC Labs America\\
% San Jose, USA \and
% Francesco Pittaluga \\
% NEC Labs America\\
% San Jose, USA \and
% Buyu Liu \\
% NEC Labs America\\
% San Jose, USA \and
% Manmohan Chandraker\\
% NEC Labs America, UC San Diego\\
% San Jose, USA

% % \and
% % Second Author\\
% % First line of institution2 address\\
% % {\tt\small secondauthor@i2.org}
% % }

\maketitle

\input{./abstract.tex}
\input{./sec_introduction.tex}
\input{./sec_related_work.tex}

\input{./sec_dac_method.tex}

\input{./sec_lane_anch_method.tex}

\input{./sec_experiments.tex}
\input{./sec_conclusion.tex}

{\small
\bibliographystyle{ieee_fullname}
\bibliography{references.bib}
}

\newpage
\setcounter{section}{0}
\setcounter{figure}{0}
\setcounter{table}{0}
% \author{}
% \maketitle
% \titlerunning{SMART Supplementary Material}
% \authorrunning{S. N N, et al.}
% \pagestyle{headings}

% \phantom{.}  %necessary to add space on top before the title
% \vspace{1cm}
% \thispagestyle{empty}
\begin{center}
{\LARGE \bf Supplementary Material}\\[1cm]    
\end{center}

\input{./sec_supp_arch}
\input{./sec_supp_poly}
\input{./sec_supp_qualitative}

\end{document}

%% file: abstract.tex
\begin{abstract}
    
    Trajectory prediction is a safety-critical tool for autonomous vehicles to plan and execute actions. Our work addresses two key challenges in trajectory prediction, learning multimodal outputs, and better predictions by imposing constraints using driving knowledge. Recent methods have achieved strong performances using Multi-Choice Learning objectives like winner-takes-all (WTA) or best-of-many. But the impact of those methods in learning diverse hypotheses is under-studied as such objectives highly depend on their initialization for diversity. As our first contribution, we propose a novel Divide-And-Conquer (DAC) approach that acts as a better initialization technique to WTA objective, resulting in diverse outputs without any spurious modes. Our second contribution is a novel trajectory prediction framework called ALAN that uses existing lane centerlines as anchors to provide trajectories constrained to the input lanes. Our framework provides multi-agent trajectory outputs in a forward pass by capturing interactions through hypercolumn descriptors and incorporating scene information in the form of rasterized images and per-agent lane anchors. Experiments on synthetic and real data show that the proposed DAC captures the data distribution better compare to other WTA family of objectives. Further, we show that our ALAN approach provides on par or better performance with SOTA methods evaluated on Nuscenes urban driving benchmark.

\end{abstract}

%% file: sec_introduction.tex
\section{Introduction}

Prediction of diverse multimodal behaviors is a critical need to proactively make safe decisions for autonomous vehicles. A major challenge lies in predicting not only the most dominant modes but also accounting for the less dominant ones that might arise sporadically. Hence, there is  need for models that can disentangle the plausible output space and provide diverse futures for any given number of samples. 
Further, a vast majority of actors execute socially acceptable maneuvers that adhere with the underlying scene structure. Predicting socially non-viable outputs can lead to unsafe planning decisions with some more dangerous than the others \cite{importance_prior_knowledge}. For example, a method that provides close enough predictions that does not follow road semantics is more dangerous compared to similar performing method that adheres to the scene structure.

\input{./figures/teaser}

Traditionally, generative models have been widely adapted to capture the uncertainties related to trajectory prediction problems \cite{DEISRE, trajectron, trajectronpp, diversitygan, sriram2020smart}. However, generative methods may suffer from mode collapse issues, which reduces their applicability for safety critical applications such as self-driving cars.
% ~\buyu{One of the contributions we claimed in SMART is the diverse predictions}. 
Recent methods \cite{mhajam, laneGCN} use Multi-Choice Learning objectives \cite{stochasticMCL} like winner-takes-all (WTA) but suffer from instability associated with network initialization \cite{ewta,rwta}. As a first contribution, we propose a Divide and Conquer (DAC) approach that provides a better initialization to the WTA objective. Our method solves issues related to spurious modes where some hypotheses are either untrained in the training process, or reach equilibrium positions that do not represent any part of the data. We show that the proposed DAC captures the data distribution better on both real and synthetic scenes with multi-modal ground truth, compared to baseline WTA objectives \cite{ewta,rwta}.

Further, trajectory prediction methods incorporate driving knowledge using scene context either in the form of rasterized images \cite{DEISRE, trajectronpp, sriram2020smart, mhajam, covernet, multipath} or by exploiting HD map data structure \cite{laneGCN, vectornet} as inputs. Usually, this information is represented as a feature given as input to the network and does not guarantee strong semantic coupling. 
Our second contribution addresses this by proposing ALAN, a novel trajectory prediction framework that uses lane centerlines as anchors to predict trajectories (Figure \ref{fig:teaser}).
% ~\buyu{Does this (centerline) actually require more information than just using pixel-level semantic maps since it requires the "lane" information? Are they comparable?}. 
Our outputs provide accurate predictions with strong semantic alignment demonstrated by FDE and OffRoadRate values and validated using our qualitative visualizations. 
% Our second contribution proposes a novel trajectory prediction framework that uses lane centerlines as anchors for predicting trajectories (Figure \ref{fig:teaser}). Our outputs provide accurate predictions with strong coupling demonstrated by FDE and OffRoadRate values and validated using our qualitative visualizations. 

Specifically, we use a single representational model \cite{sriram2020smart} for multi-agent inputs and encode interactions through novel use of hypercolumn descriptors \cite{pixelnet} that extracts information from features at multiple scales. Moreover, we transform the prediction problem to normal-tangential ($nt$) coordinates with respect to input lanes. This is critical in order to use lane centerlines as anchors. Further, we regularize anchor outputs through auxiliary $xy$ predictions to make them less susceptible to bad anchors and rely on agent dynamics. Finally, we rank our predictions through an Inverse Optimal Control based ranking module \cite{DEISRE}.

In summary, our contributions are the following:\begin{itemize}\itemsep-2pt
    \item A novel Divide and Conquer approach as a better initialization to WTA objective that captures data distribution without any spurious modes. 
    \item A new anchor based trajectory prediction framework called ALAN that uses existing centerlines as anchors to provide context-aware outputs with strong semantic coupling.
    \item Strong empirical performance on the Nuscenes urban driving benchmark.
\end{itemize}

%% file: figures/teaser.tex
\begin{figure}[t]
    \centering
    \includegraphics[width=1.0\linewidth]{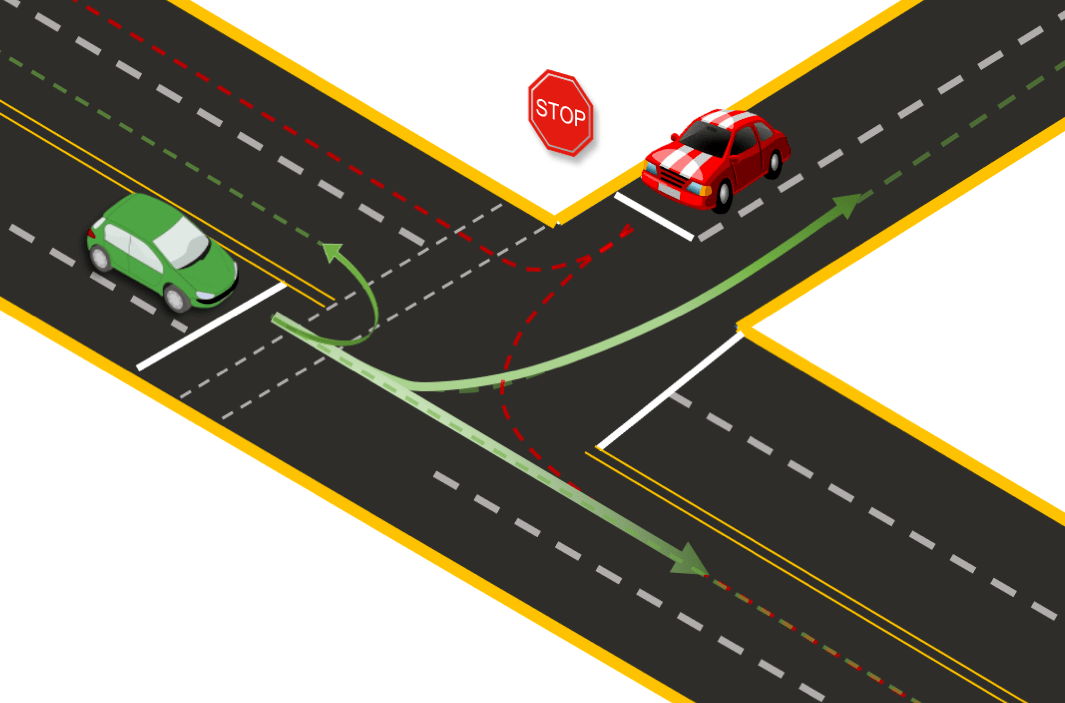}
    \caption{Depicts trajectory prediction problem in an intersection scenario with possible lane anchors for agents shown as coloured dashed lines.}
    \label{fig:teaser}
\end{figure}

%% file: sec_related_work.tex
\section{Related Work}

\textbf{Multi-Choice Learning:} Multi-modal predictions have been realized in different domains through Multi-choice learning (MCL) \cite{mcl2012,dey2015predicting, stochasticMCL} objectives in the past.  Several works have shown use cases of MCL to provide diverse hypotheses in classification \cite{stochasticMCL, rwta}, segmentation \cite{stochasticMCL, rwta}, captioning \cite{stochasticMCL}, pose estimation \cite{rwta}, image synthesis \cite{chen2017photographic} and trajectory proposals \cite{sriramIV}. Convergence issue related to WTA objectives have been shown in \cite{rwta,ewta}. Following this work, \cite{rwta} proposed a relaxed winner-takes-all objective (RWTA) to solve the convergence problem but this method itself suffers from the problem of hypotheses incorrectly capturing the data distribution. \cite{ewta} proposed an evolving winner-takes-all (EWTA) loss that captures the distribution better compared to \cite{rwta}. Despite the aforementioned improvements, these methods still can't capture the data distribution accurately due to spurious modes at equilibrium or hypotheses untrained during the training process. Alternatively, we propose a Divide and Conquer approach where we exponentially increase the effective number of outputs during training with set of hypotheses capturing some part of the data at every stage.

\textbf{Forecasting Methods:}
% \subsubsection{}
% ~\buyu{Related work seems to be too short while the intro is quite long.}
The future trajectory prediction has been investigated broadly in the literature using both classical \cite{classic_reciprocal, classic_exploiting, calssic_activity} and deep learning based methods \cite{Social_gan, det_alahi2016social, sriram2020smart}. Deterministic models \cite{det_alahi2016social,det_luo2018fast,det_carnet} predict most likely  trajectory for each agent in the scene while neglecting the uncertainties inherited in the trajectory prediction problem. To capture the uncertainties and create diverse trajectory predictions, stochastic methods have been proposed which encode possible modes of future trajectories through sampling random variables. Non-parametric deep generative models such as Conditional Variational Autoencoder (CVAE) \cite{DEISRE,best_of_many,trajectron,Rules_of_the_road,sriram2020smart} and Generative Adversarial Networks (GANs) \cite{Social_bigat,Social_gan,Sophie} have been widely used in this domain. However, these methods fail to capture all underlying modes due to imbalance in the latent distribution \cite{yuan2019diverse}. Recent methods predict a fixed set of diverse trajectories \cite{mhajam, laneGCN} for the same input context. Our method uses a similar approach to predict a set of $M$ hypothesis.

% However, these methods are usually susceptible to mode-collapse and they usually are unable to capture all underlying modes of the future trajectories distribution.
% \textbf{HD Maps And Anchor Based Prediction:}
\textbf{Representation:}
% \sriram{Need to talk about different type of prediction methods, xy, classific, nt. Acknowledge Argoverse in using nt but only basic NN and Lstm. Other methods have not explored further}
HD map rasterization have been widely used in the literature to encode and process map information by neural networks \cite{best_of_many, map_tensor_fusion, map_motion, map_implicit, sriram2020smart}. Some methods \cite{srikanth2019infer, marchetti2020mantra} construct top view map using semantics and depth information from perspective images. Some \cite{Dsdnet, map_implicit} use a combination rasterzied HD maps and sensor information. Several recent works \cite{laneGCN,vectornet} utilize map information directly by representing the vectorized map data as a graph data structure. Our work uses a hybrid map input combining both rasterized map and vectorized lane data provided as input for every agent at its location on the spatial grid \cite{sriram2020smart}.

\textbf{Trajectory Prediction:} Traditionally, several works \cite{sriram2020smart, laneGCN, vectornet, mhajam} formulate trajectory prediction problem as a regression over cartesian coordinates. \cite{srikanth2019infer} poses it as a classification of future locations over a spatial grid. Chang et. al \cite{argoverse} use a normal-tangential coordinate similar to ours but is only limited to classical nearest neighbor and vanilla LSTM \cite{lstm} approaches. Related to our work, some methods tackle the multi-modality problem by quantizing the output space into several predefined diverse anchors and then reformulating the original trajectory problem into sequential anchor classification (selection) and offset regression sub-problems \cite{TNT,covernet, multipath,Dsdnet}. However, Anchors usually are  pre-clustered into a fixed set as a priori or are calculated in real-time based on kinematic heuristics \cite{TNT}. Hence, the process of creating anchors may add computational complexity in the inference time, also it could be highly scenario dependent and hard to generalize. In contrast, our method uses HD map centerline information as anchors which is consistent for diverse scenarios and also readily available at inference. 

\input{./figures/dac_toy}

%% file: figures/dac_toy.tex
\begin{figure*}[t]
    \centering
    \begin{subfigure}{0.24\linewidth}
        \includegraphics[width=\linewidth]{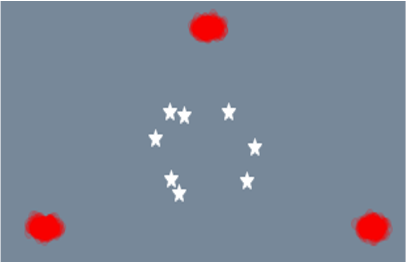}
        \caption{Init}
        \label{fig:toy_dac_init}
    \end{subfigure}
    \begin{subfigure}{0.24\linewidth}
        \includegraphics[width=\linewidth]{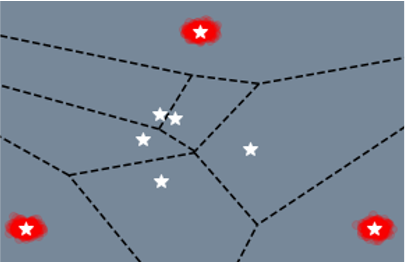}
        \caption{WTA}
        \label{fig:toy_dac_wta}
    \end{subfigure}
    \begin{subfigure}{0.24\linewidth}
        \includegraphics[width=\linewidth]{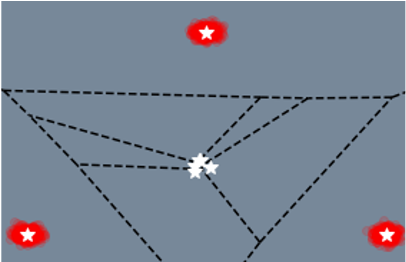}
        \caption{RWTA}
        \label{fig:toy_dac_rwta}
    \end{subfigure}
    \begin{subfigure}{0.24\linewidth}
        \includegraphics[width=\linewidth]{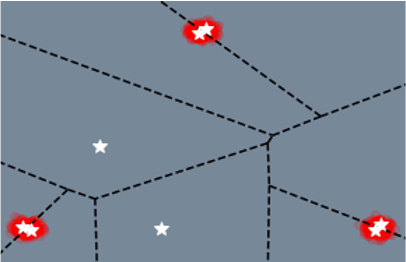}
        \caption{EWTA}
        \label{fig:toy_dac_ewta}
    \end{subfigure}

    \begin{subfigure}{0.24\linewidth}
        \includegraphics[width=\linewidth]{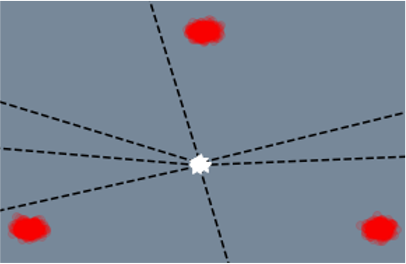}
        \caption{DAC - Depth 1}
        \label{fig:toy_dac_d1}
    \end{subfigure}
    \begin{subfigure}{0.24\linewidth}
        \includegraphics[width=\linewidth]{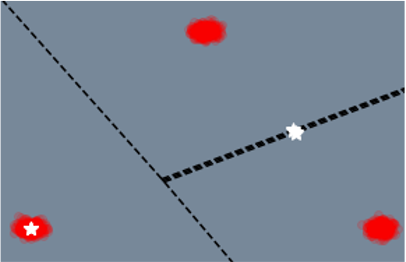}
        \caption{DAC - Depth 2}
        \label{fig:toy_dac_d2}
    \end{subfigure}
    \begin{subfigure}{0.24\linewidth}
        \includegraphics[width=\linewidth]{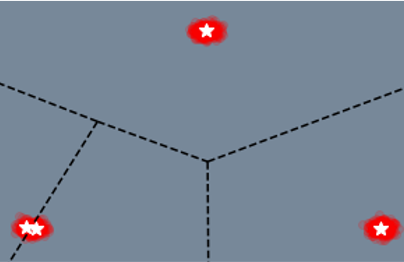}
        \caption{DAC - Depth 3}
        \label{fig:toy_dac_d3}
    \end{subfigure}
    \begin{subfigure}{0.24\linewidth}
        \includegraphics[width=\linewidth]{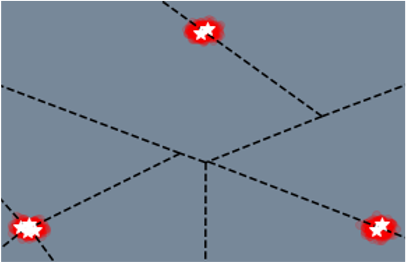}
        \caption{DAC - Depth 4}
        \label{fig:toy_dac_d4}
    \end{subfigure}
    \caption{\small Toy example comparing different versions of winner-takes-all and enclosed voronoi regions of their predicted hypotheses. The toy data is shown in red and the hypotheses are shown in white. With Depth=1 for DAC, it contains a single set with $M$ hypotheses, thus all hypotheses are penalized to match the data and reach the equilibrium. As the depth increases the number of sets in the list grows exponentially as every set is broken down into halves ($e \xrightarrow[]{} f \xrightarrow{} g \xrightarrow{} h$).
    % and the set containing argmin output in the list is penalized to match the ground truth. 
    Since we show the same ground truths to all the hypotheses in a set, they reach the same equilibrium position forming centroidal voronoi tessellation with number of outputs effectively equal to the number of sets in the list ($e \xrightarrow{} 1, f \xrightarrow{} 2, g \xrightarrow{} 4, h \xrightarrow{} 8$). 
    % With an increase in depth from f to g, the hypotheses are split into halves with two reaching the equilibrium on the left and the other two on the right. 
    % This makes every hypothesis in the set reach the same equilibrium position forming centroidal voronoi tessellation with number of outputs effectively equal to the number of sets in the list. 
    In the final stage (h), every set contains one hypothesis resembling a WTA objective. 
    % As we reach the leaf nodes, every set will contain one hypothesis forming a winner-takes-all objective. 
    % With depth=1 in DAC, all $M$ hypothesis is penalized to match the data, thus reaching the equilibrium. As the depth increases, the number of elements 
    % With depth of the tree in DAC=1, the set contains a single element penalizing all $M$ hypothesis to match the data, thus reaching the equilibrium. As the depth of the tree increases number of elements in the set grows exponentially as every element is broken down into halves 
    % With number of sets in the DAC=1, it penalizes all the hypothesis to match the data distribution. Thus 
    In comparison to DAC, other WTA objectives model the data distribution incorrectly since some Voronoi regions do not capture any part of the data, resulting in spurious modes. 
    % As seen some voronoi regions of other wta objectives do not capture any part of the data 
    % Illustrative toy example comparing different versions of winner-takes-all and enclosed voronoi regions of their predicted hypothesis.
    }
    \label{fig:toy_dac}
\end{figure*}

%% file: sec_dac_method.tex
\section{Divide and Conquer}

In this section, we provide detailed description of our method in training Multi-Hypothesis prediction networks where our approach acts as an initialization technique for winner-takes-all~\cite{stochasticMCL} objective. Let $\mathcal{X}$ denote the vector space of inputs and $\mathcal{Y}$ denotes the vector space of output variables. Let $\mathcal{D} = \{(x_i, y_i), ..., (x_N, y_N)\}$ be a set of $N$ training tuples and $p(x,y) = p(y|x)p(x)$ be the joint probability density. % where $p(y|x)$ is the conditional probability for label $y$ given label $x$. 
Our goal is to learn a function $f_\theta : \mathcal{X}\xrightarrow{}\mathcal{Y}^M$ that maps every input in $\mathcal{X}$ to a set of $M$ hypotheses. Mathematically, we define:
\begin{equation}
    f_\theta(x) =  (f_\theta^1(x), ..., f_\theta^M(x)).
\end{equation}

As shown by Rupprecht et al.~\cite{rwta}, winner-takes-all objective minimizes the loss with the closest of $M$ hypotheses:
\begin{equation}
    \int_{\mathcal{X}}\sum_{j=1}^M \int_{\mathcal{Y}_{j(x)}} \mathcal{L}(f_\theta^j(x),y)p(x,y) dy dx, 
\end{equation}
where $\mathcal{Y}_j$ is the Voronoi tessellation of label space with $\mathcal{Y} = \cup_{j=1}^M\mathcal{Y}_j$. This objective leads to Centroidal Voronoi tessellation \cite{centroidalVoronoi} of outputs where each hypothesis minimizes to the probabilistic mass centroid of the Voronoi label space $\mathcal{Y}_j$ enclosed by it. In practice, to obtain diverse hypotheses WTA objective can be written as a meta loss \cite{ewta,rwta,stochasticMCL,mcl2012},

\begin{equation}
    \mathcal{L}_{WTA} = \sum_{k=1}^K \delta_k(k == \argmin_i \mathcal{L}(f_\theta^i)) \mathcal{L}(f_\theta^k(x),y),
    \label{eq:wta}
\end{equation}
where $\delta (\cdot)$  is the Kronecker delta function with value 1 when condition is True and 0 otherwise.  
% \MC{Rather put the full expression for $\delta_k$ in \eqref{eq:wta} and can write $\delta (\cdot)$ here.}

\vspace{-0.3cm}
\paragraph{Initialization difficulties for WTA}
As mentioned by Makansi et al.~\cite{ewta} Equation \ref{eq:wta} can be compared to EM algorithm and K-means clustering where they depend mainly on initialization for optimal convergence. As shown in \ref{fig:toy_dac_wta} this makes training process very brittle as the Voronoi region of only few hypotheses encloses the data distribution, leaving most of the hypotheses untrained due to winner-takes-all objective. The alternative solution proposed by Ruppercht et al.~\cite{rwta} to solve the convergence problem by assigning $\epsilon$ weight to the non-winners does not work as every ground truth associates with atmost one hypothesis making other non-winners to reach the equilibrium as shown in \ref{fig:toy_dac_rwta}. Makansi et. al, \cite{ewta} then proposed evolving winner-takes-all (EWTA) objective where they update top $k$ winners. The $k$ varies starting from $k=M$ to $k=1$ leading to winner takes all objective in training process. This method captures the data distribution better compared to RWTA and WTA but still produces hypothesis with incorrect modes as shown in the Figure \ref{fig:toy_dac_ewta}.

\vspace{-0.3cm}
\paragraph{DAC for diverse non-spurious modes}
We propose an novel initialization technique called Divide and Conquer that alleviates the problem of spurious modes, leaving the Voronoi region of every output hypothesis to capture some part of the data, as shown in Figure \ref{fig:toy_dac_d4}. We divide $M$ hypotheses into $k$ sets and update the set with argmin outputs to match the ground truth. The value of $k$ starts with $1$ and increases exponentially as every set is broken down into two halves as we progress through the training. This creates a binary tree with the depth of the tree dependent on the number of output hypotheses $M$. Algorithm \ref{alg:DAC} shows pseudo-code of the proposed Divide and Conquer technique. Here \texttt{depth} specifies the maximum depth that can be reached in the current training stage and we define \texttt{list} as variable containing set of hypotheses at any stage in the training. Further, we define newly formed sets from $k^{th}$ set as $set_{k1}$ and  $set_{k2}$. Set from the \texttt{list} that produces argmin output is denoted as \texttt{mSet}. Finally we take mean loss of all hypotheses in \texttt{mSet} to get $\mathcal{L}_{DAC}$.
% \MC{There are some terms like set or list here that are not clear. Use precise notation to describe here and use it in Algorithm 1 too.} 

% \MC{Improve notation in Alg 1. I don't think it is actually correct the way it is currently. $k$ is the depth level in step 3, but then you have $set_k$. Also $k1$, $k2$ are undefined. Don't use min as a variable name and as an operation in line 7. $min_{set}$ definition reads like the set of all $set_k$. Not sure what $set = \{ loss \}$ means.}

From Figure \ref{fig:toy_dac_d1}, with $k=1$ and \texttt{list} containing a single set, all $M$ hypotheses reach towards the equilibrium. 
% \MC{``$k$ containing a single set'': I guess you mean ``with $k = 1$ and the list containing a single set''.}
As the number of sets in the list increases from \ref{fig:toy_dac_d1} to \ref{fig:toy_dac_d2} the hypotheses divide the distribution space based on the Voronoi region to capture different parts of the data. 
% \textcolor{red}{Ramin: could you take a look on the next sentence and revise it if it is necessary?  } With an increase in depth leading to more sets the equilibrium position for all the hypotheses changes as newly formed sets only look at the subpart of the data from their previous stage. 
The effective number of outputs grows at every stage, with the data captured by the $k^{th}$ set in the previous stage split across two newly formed sets in the next stage.
Finally, as we reach the leaf nodes, every set contains one hypothesis leading to a winner-takes-all objective similar to Equation \ref{eq:wta}.

DAC starts with all hypotheses fitting the whole data and at every stage DAC ensures some data to be enclosed in the Voronoi space. During split, hypotheses divide the data enclosed within their Voronoi space to reach new equilibrium. Although, DAC does not guarantee equal number of hypotheses capturing different modes of the data it ensures convergence. Further we would like to note that DAC does not have any significant computational complexity as only dividing into sets and min calculations are involved. In Section \ref{sec:exp}, we show benefits of DAC in capturing multimodal distributions better, producing diverse set of hypotheses compared to other WTA objectives.

\begin{algorithm}
\caption{Divide and Conquer technique}
\begin{algorithmic}[1]
% \State MaxDepth = \texttt{int(log(M)/log(2))}
\Procedure{DAC}{\texttt{loss, depth}}
    \State $set_1 =$ \texttt{$\{$loss$\}$} \Comment{All M hypotheses}
    \State \texttt{list} = [$set_1$]
    \For{\texttt{$i \gets 2$ {\bf to} depth}}
        \For{\texttt{$set_k \in \texttt{list}$}}
            \State \texttt{$\sslash$ Divide $set_k$ into halves}
            \State \texttt{list} $+= [\{set_{k1}\}, \{set_{k2}\}]$
            % \State \texttt{$set_k = \{set_{k1}, set_{k2}\}$}
        \EndFor
    \EndFor
    % \State $\min_{set} = \{$\parbox[t]{0.7\linewidth}{$set_k : \min(set_k) < \min(set_j);\forall j \in \{1..len(set)\}, j \not= k\}$}
    \State $\texttt{mSet} = \{$\parbox[t]{0.7\linewidth}{$set_k : \min(set_k) < \min(set_j);\forall j \in \{1..len(\texttt{list})\}, j \not= k\}$}
    % \State $\mathcal{L}_{DAC} = mean(\min_{set})$
    \State $\mathcal{L}_{DAC} = mean(\texttt{mSet})$
    \State \textbf{return} $\mathcal{L}_{DAC}$
\EndProcedure
\end{algorithmic}
\label{alg:DAC}
\end{algorithm}

% \section{SMART++ approach}

% In this section we will introduce a single representation model to predict trajectories using Lane Anchors for multiple agents that are semantically aligned trajectories. 

% This section provide a detailed description of SMART++ approach. We first describe the problem statement and then provide descriptions to various 
% In this section we will introduce a single representation model to predict trajectories using Lane Anchors for multiple agents that are semantically aligned and . We formulate the trajectory prediction problem as one-shot regression of vehicle coordinates across timesteps. We will provide a detailed description of our method below. 

% \subsection{Notation:}

%% file: sec_lane_anch_method.tex
\section{Trajectory Prediction with Lane Anchors}
\input{./figures/pipeline}

In this section, we introduce a single representation model called ALAN that produces lane aware trajectories for multiple agents in a forward pass. We formulate the problem as one shot regression of diverse hypotheses across time steps. We now describe our method in detail. 

\subsection{Problem Statement} 
Our method takes scene context input in two forms: a) rasterized birds-eye-view (BEV) representation of the scene denoted as ${\bf I}$ of size $H \times W \times 3$ and b) per-agent lane centerline information as anchors. We define lane anchors ${\bf L} = \{ L_1,...L_p\}$ as a sequence of $p$ points with coordinates $L_p = (x,y)$ in the BEV frame of reference. We denote ${\bf X}_i = \{X_i^1, ... X_i^{T}\}$ as trajectory coordinates containing past and future observations of the agent $i$ in Cartesian form, where $X_i^t = (x_i^t, y_i^t)$.  
% ~\buyu{We have multiple definitions for x and y. Better to change to other characters.} 
For every agent $i$, we identify a set of candidate lanes that the vehicle may take based on trajectory information like closest distance, yaw alignment and other parameters (see supplementary). We denote this as a set of plausible lane centerlines $\mathcal{A} = \{ {\bf L}_1, ...,{\bf L}_k\}$, where $k$ represents total number of lane centerlines along which the vehicle may possibly travel. We then define vehicle trajectories ${\bf X}_i$ along these centerlines in a 2d curvilinear normal-tangential ($nt$) coordinate frame.
% ~\buyu{Missing i when discussing i-th object? And we already introduced $X_i$ above.}\sriram{Done}. 
We denote ${\bf N}_{i,k} = \{N_{i,k}^1, ..., N_{i,k}^{T}\}$ as the $nt$ coordinates for the agent $i$ along the centerline ${\bf L}_k$, where $N_{i,k}^t = (n_{i,k}^t, l_{i,k}^t)$ denotes normal and longitudinal distance to the closest point along the lane.
% \MC{The symbol $t$ would be expected to be the second coordinate of the $nt$ system, where $l$ is used here with $t$ for time step.}
% ~\buyu{to the shortest point in centerline $L_k$? Also, it is unusual to put m on the top left of N, better to have them all on the right hand side.}\sriram{Done}
Use of $nt$ coordinates is crucial to capture complex road topologies and associated dynamics to provide predictions that are semantically aligned and has been studied in our experiments (Section \ref{sec:exp}). 
% \MC{Write a sentence here to say that the choice of $nt$ coordinates is crucial (can give a hand-wavy explanation: it better captures the relationship between vehicle dynamics and road topology in complex scenes) and say this is studied in the experiments.}

We then define trajectory prediction problem as the task of predicting $^{nt}{\bf Y}_{i,k} = \{N_{i,k}^{t_{obs}}, ..., N_{i,k}^{T}\}$ for the given lane anchor ${\bf L}_k$ provided as input to the network. We follow an input representation similar to \cite{sriram2020smart}, where we encode agent specific information at their respective $X_i^{t_{obs}}$ locations on the spatial grid. Finally, to get trajectories in BEV frame of reference we convert our output predictions to cartesian coordinates based on the anchor ${\bf L}_{i,k}$ given as input to the network.
% ~\buyu{We should note that BEV (bird-eye-view) and top view are exchangeable in our paper.} 
% Finally, we calculate our output trajectories in cartesian form based on the anchor ${\bf L}_k$ given as input to the network. 

\subsection{ALAN Framework for Trajectory Prediction}
An overview of our framework is shown in Figure \ref{fig:pipeline}. Our method consists of five major components: a) a centerline encoder b) a past trajectory encoder c) a multi-agent convolutional interaction encoder d) hypercolumn \cite{pixelnet} trajectory decoder and e) an Inverse Optimal Control (IOC) based ranking module \cite{DEISRE}.

\vspace{-0.3cm} 
\paragraph{Centerline Encoder:} We encode our input lane information ${\bf L}_{i,k}$ for every agent through a series of 1D convolutions to produce an embedded vector ${\bf C}_{i,k} = \mathcal{C}_{enc}({\bf L}_{i,k}) $ for every agent in the scene. 

\vspace{-0.3cm} 
\paragraph{Past Trajectory Encoder:}  Apart from $nt$ coordinates ${\bf N}_{i,k}$ for the lane anchor, we provide additional ${\bf X}_i$ as input to the past encoder. We first embed the temporal inputs through a $MLP$ and then pass it through a LSTM\cite{lstm} network to provide a past state vector ${\bf h}_i^{t_{obs}}$. Formally,

\begin{equation}
    {\bf s}_i^{t} = MLP(X_i^t, N_{i,k}^t)
\end{equation}
\begin{equation}
    {\bf h}_i^{t_{obs}} = LSTM({\bf s}_i^{1..t_{obs}})
\end{equation}

\vspace{-0.3cm} 
\paragraph{Multi-Agent Convolutional Encoder:} We realize multi-agent prediction of trajectories in a forward pass through a convolutional encoder module \cite{sriram2020smart}. 
% \MC{If this is like SMART, then cite it here.} 
First, we encode agent specific information ${\bf C}_{i,k}, {\bf h}_i^{t_{obs}}$ at their respective locations $X_i^{t_{obs}}$ in the BEV spatial grid. This produces a scene state map ${\bf S}$ of size $H \times W \times 128$ containing information of every agent in the scene. We then pass this through a convolutional encoder along with the rasterized BEV map ${\bf I}$ to produce activations at various feature scales. In order to calculate feature vectors of each individual agent, we adapt a technique from Bansal et al.~\cite{pixelnet} to extract hypercolumn descriptors ${\bf D}_i$ from their locations. The hypercolumn descriptor contains features extracted at various scales by bi-linearly interpolating $X_i^{t_{obs}}$ for different feature dimensions. Thus,
\begin{equation}
    {\bf D}_i = [c_1(X_i^t), ..., c_k(X_i^t)],
\end{equation}
where $c_k$ is the feature extracted at $k^{th}$ layer by bilinearly interpolating the input location to the given dimension. The intuition is to capture interactions at different scales when higher convolutional layers capturing the global context and low-level features retaining the nearby interactions. In Section \ref{sec:exp}, we show using hypercolumn descriptors in trajectory prediction task can be beneficial compared to just using global context vectors.

% \MC{Should claim novelty for hypercolumn use in the trajectory prediction task.}

\vspace{-0.3cm} 
\paragraph{Hypercolumn Trajectory Decoder:} The hypercolumn descriptor ${\bf D}_i$ of every agent is then fed through a decoder containing a series of 1x1 convolutions to output $M$ hypotheses at once. Here we investigate two variants of ALAN prediction. ALAN-$nt$ where we predict $nt$ trajectories $^{nt}{\bf \hat Y}_i$ in the direction of the lane and ALAN-$ntxy$ which also provides an auxiliary $xy$ predictions $^{xy}{\bf \hat Y}_i$. Linear values in $nt$ can correspond to trajectories of higher degrees based on the input anchor. Moreover, two trajectories having same $nt$ values can have completely different dynamics. Thus we make use of the auxiliary predictions to regularize anchor based outputs to make the network aware of agent dynamics and less susceptible to bad anchors. The $M$ hypotheses predicted from our network is given as:
% ALAN-NT where we predict trajectories in the direction of the lane ($nt$) and ALAN-NTXY where we provide auxiliary ${\bf X}_i^{t_{obs}..T}$ predictions in $xy$. Linear values in $^k{\bf N}_i$ can correspond to trajectories of higher degrees based on the input anchor. Moreover, two trajectories having same ${\bf N}$ coordinates can correspond to completely different dynamics. Thus we make use of the auxiliary predictions to regularize anchor based outputs to make the network aware of agent dynamics and less susceptible to bad anchors. Thus output predictions from our network can be given as,
\begin{equation}
    ^{nt}{\bf \hat Y}_i, ^{xy}{\bf \hat Y}_i = CNN_{1*1}({\bf D}_i),
\end{equation}
\begin{equation}
    ^{nt}{\bf \hat Y}_i = \{^{nt}{\hat Y}_{i,1}, ^{nt}{\hat Y}_{i,2}..., ^{nt}{\hat Y}_{i,M}\},
\end{equation}
\begin{equation}
    ^{xy}{\bf \hat Y}_i = \{^{xy}{\hat Y}_{i,1}, ^{xy}{\hat Y}_{i,2}..., ^{xy}{\hat Y}_{i,M}\}.
\end{equation}

\vspace{-0.3cm} 
\paragraph{Ranking Module:} We use the technique from Lee et al.~\cite{DEISRE} to generate scores $^s{\bf Y}_i = \{^sY_{i,1}, ^sY_{i,2},..., ^sY_{i,M}\} $ for the $M$ output hypotheses. It measures the goodness $^sY_{i,k}$ of predicted hypotheses by assigning rewards that maximizes towards their goal\cite{sutton2018reinforcement}.
% It learns to assign reward to each predicted hypothesis and measures their goodness $^sY_{i,k}$.
% \MC{A sentence to motivate why this is useful.} 
The module uses predictions $^{nt}{\bf \hat Y}^i$ to obtain the target distribution ${q}$,  where $q = \texttt{softmax}(-d(^{nt}Y_i, ^{nt}{\hat Y}_i))$ and $d$ being the L2 distance between the ground truth and predicted outputs. Thus, the score loss is given as $\mathcal{L}_{score} =$ Cross-Entropy$(^s{\bf Y}_i, {\bf q})$.

% ~\buyu{Should be in different level w.r.t. the above mentioned modules} 
\subsection{Learning}
We supervise the network outputs $\{^{nt}{\bf \hat Y}_i, ^{xy}{\bf \hat Y}_i\}$ as the L2 distance with their respective ground truth labels $^{nt}{\bf Y}$ for the input lane anchor ${\bf L}_k$ and $^{xy}{\bf Y}$. We use the proposed Divide and Conquer technique to train our Multi-Hypothesis prediction network. Hence, the reconstruction loss for both primary and auxiliary predictions is given by:
\begin{equation}
    ^{nt}\mathcal{L}_{DAC} = \texttt{DAC}(^{nt}{\bf \hat Y}_i),
\end{equation}
\begin{equation}
    ^{xy}\mathcal{L}_{DAC} = \texttt{DAC}(^{xy}{\bf \hat Y}_i).
\end{equation}

Additionally, we penalize our anchor based predictions based on $^{xy}{\bf \hat Y}_i$ by transforming the predictions to $nt$ coordinates $^{xy}{\bf \hat Y}_i^{nt}$ along the input lane. We also add the regularization other way to penalize $^{xy}{\bf \hat Y}_i$ predictions based on the anchor outputs $^{nt}{\bf \hat Y}_i$ by converting them to $xy$ coordinates $^{nt}{\bf \hat Y}_i^{xy}$. We add the regularization as L2 distance between the converted primary and auxiliary predictions for all hypotheses:
\begin{equation}
    ^{nt}\mathcal{L}_{xy} = L2(^{nt}{\bf \hat Y}_i, ^{xy}{\bf \hat Y}_i^{nt}),
\end{equation}
\begin{equation}
    ^{xy}\mathcal{L}_{nt} = L2(^{xy}{\bf \hat Y}_i, ^{nt}{\bf \hat Y}_i^{xy}).
\end{equation}

The total learning objective for the network to minimize can be given by,
\begin{equation}
    \begin{split}
        \mathcal{L} = & ^{nt}\mathcal{L}_{DAC} + ^{xy}\mathcal{L}_{DAC} \\ & + \lambda_1  ^{nt}\mathcal{L}_{xy} + \lambda_2  ^{xy}\mathcal{L}_{nt} + \mathcal{L}_{score}.
    \end{split}
\end{equation}

% \begin{equation}
%     {\bf S} = \phi(^k{\bf C}_i \oplus {\bf h}_i^{t_{obs}})
% \end{equation}
% \begin{equation}
%     {\bf \psi} = CNN({\bf S})
% \end{equation}

%% file: figures/pipeline.tex
\begin{figure*}[ht]
    \centering
    \includegraphics[width=1.0\textwidth]{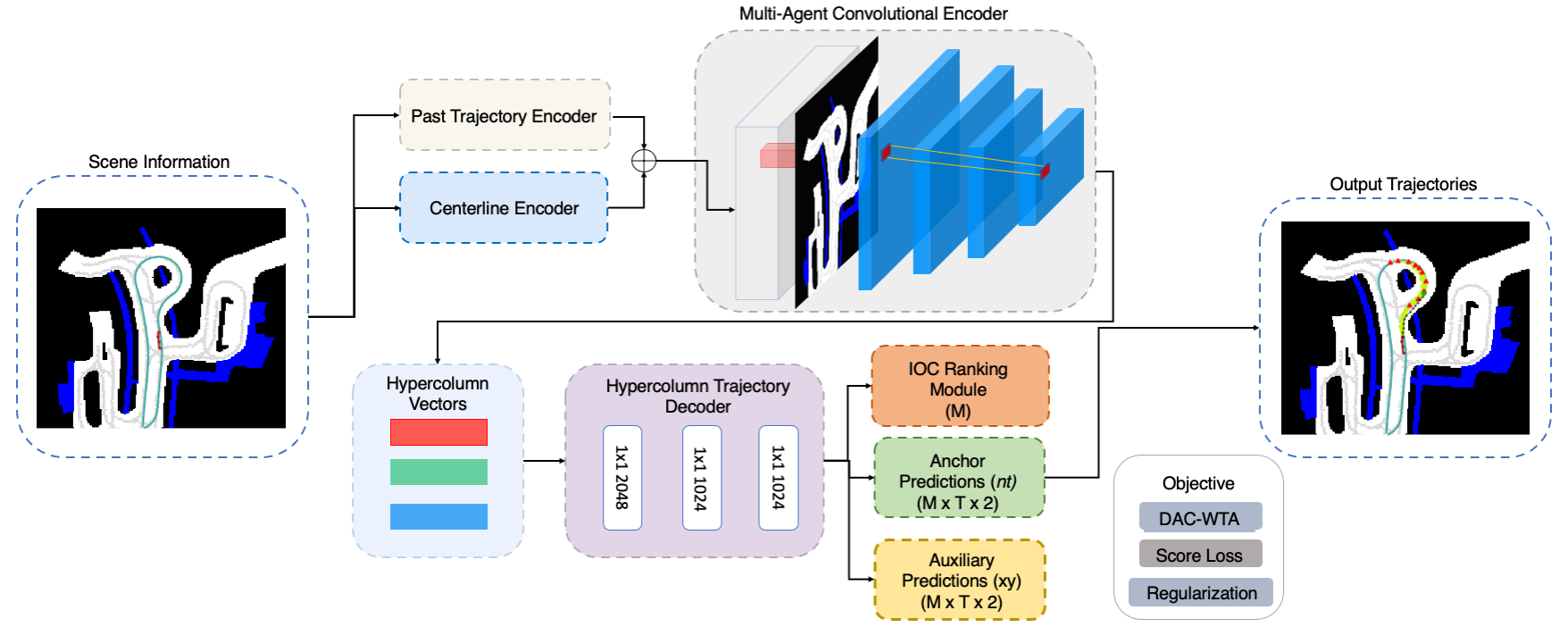}
    \caption{Overview of our proposed ALAN approach. The method takes in past trajectory along with lane anchor and BEV map as input to provide multi-hypothesis predictions for all agents at once.}
    \label{fig:pipeline}
\end{figure*}

%% file: sec_experiments.tex
\section{Experiments}\label{sec:exp}
\input{./figures/cpi_img}

We first evaluate our proposed Divide and Conquer technique on the synthetic Car Pedestrian dataset\cite{ewta}. Further, we show evaluations of DAC and the proposed anchor based prediction technique on Nuscenes\cite{nuscenes2019} prediction dataset.
% We evaluate our proposed Divide and Conquer technique on both synthetic and real world datasets. Further, we show SOTA performance of our ALAN on Nuscenes Trajectory Forecasting Benchmark. 

% \subsection{Divide and Conquer}
\subsection{Car Pedestrian Dataset}
% We utilize synthetic Car Pedestrian Dataset (CPI) proposed in \cite{ewta} to evaluate our DAC approach.
Unlike real world settings where only a single outcome is observed, CPI 
% \ramin{is not it necessary to explain CPI?} 
dataset consists of interacting agents with multi-modal ground truths. We aim to evaluate how well our multi-hypothesis predictions capture the true distribution of samples in the test set. We use a similar training strategy from \cite{ewta} using a ResNet-18 \cite{resnet18} encoder backbone where we train a two-stage mixture density network \cite{Bishop1994MixtureDN}. The first stage takes past observations of the car and pedestrian as the inputs and predicts $k$ output hypotheses containing future goals of both actors after $\Delta t$ timestep. We train the first stage using different variants of the winner-takes-all loss function. The second stage then fits a mixture distribution with $M$ modes over the hypothesis by predicting soft-assignments for the outputs. We refer readers to Equations 7, 8 and 9 from \cite{ewta} for more details about calculating the parameters for the mixture distribution. We use evaluation metrics such oracle error (FDE) and Earth Mover's Distance (EMD) used in \cite{ewta}.

% Unlike real world settings where only a single outcome is observed, CPI \textcolor{red}{Ramin: is not it necessary to explain CPI?} dataset consists of interacting agents with multi-modal ground truths. We aim to analyse how well our multi-hypothesis predictions capture the true distribution of samples in the test set. We use a similar training strategy from \cite{ewta} using a ResNet-18 \cite{resnet18} encoder backbone where we train a two-stage mixture density network \cite{Bishop1994MixtureDN}. The first stage takes past observations of the Car and Pedestrian as input and predicts $k$ output hypothesis containing future goals of both actors after $\Delta t$ timestep. We train the first stage using different variants of the winner-takes-all  objective. The second stage then fits a mixture distribution with $M$ modes over the hypothesis by predicting soft-assignments for the outputs. We refer the reader to Equations 7,8,9 from \cite{ewta} on how we calculate the parameters for the mixture distribution. We use evaluation metrics such oracle error (FDE) and Earth Mover's Distance (EMD) used in \cite{ewta}. 
% \sriram{Can make it more detailed if needed}

\textbf{Oracle error (FDE)} measures the diversity of our outputs predictions by choosing the closest hypothesis with the ground truth.

\textbf{EMD distance} quantifies the amount of probability mass that has to be moved from the predicted distribution to match the true distribution. 

From Table \ref{tab:cpi_table} it can be inferred that the proposed DAC method outperforms the other variants of WTA objective showing that DAC captures the data distribution better compared to EWTA, RWTA and WTA. This can also been seen in Figure \ref{fig:cpi_img} where network trained with DAC objective captures the ground truth distribution of actors better compared to other variants. The average EMD of the proposed DAC is significantly better than WTA and comparable to EWTA and RWTA objective. DAC better captures goals for the cars that spread across compared to pedestrian goals. Moreover, as shown by Table \ref{tab:cpi_table}, the average oracle error (FDE) for the DAC method is significantly lower compared to other variants confirming that DAC WTA produces diverse hypotheses. 

\input{./tables/cpi_table.tex}
\input{./figures/nu_img}
\input{./tables/nuscenes_benchmark.tex}

\input{./tables/nuscenes_ablation.tex}

\subsection{Nuscenes Dataset}
Nuscenes\cite{nuscenes2019} contains a large collection of complex road scenarios from cities of Boston and Singapore.
% Nuscenes\cite{nuscenes2019} contains a large collection of complex road scenarios collected from cities of Boston and Singapore containing 1000 driving scenes. 
Approximately 40k instances were extracted for the prediction dataset. It contains challenging sequences such as ones with U-turns and complex road layouts. 
% We first evaluate our proposed ALAN predictions on this and also show ablation of the proposed DAC with other WTA variants.

\subsubsection{Baselines}\vspace{-0.1cm}
We show comparisons of our ALAN predictions with several baseline methods evaluated on Nuscenes benchmark. MTP \cite{mtp} uses rasterized image as input to predict trajectories. CoverNet \cite{covernet} uses fixed set of trajectories to solve the prediction as a classification over the trajectory set. Multipath \cite{multipath} is the closest baseline that uses time parameterized anchor trajectories obtained from the train set and formulates the problem as regression of offset values with respect to their anchor heads. MHA\_JAM \cite{mhajam} is recent method that uses joint agent-map representation to produce outputs with multi-head attentions. Trajectron++ \cite{trajectronpp} is graph recurrent model that predicts trajectories incorporating agent dynamics and semantics. We utilize the numbers for \cite{mtp} and \cite{multipath} from \cite{mhajam}. 

\subsubsection{Metrics}
We use standard evaluation metrics such as Average Displacement Error (mADE$_M$) and Final Displacement Error (mFDE$_M$). Further, we compute miss rate (Miss$_{d,M}$) of top $M$ likely trajectories with the GT. A set of predictions is considered to be a miss if there's no hypothesis across predictions having maximum displacement point less than the threshold $d$. OffRoadRate computes percentage of output trajectories that fall outside the drivable region. We use the example API provided by Nuscenes to compute our metrics.

% mADE$_k$ computes the minimum average displacement of the trajectory across hypothesis with that of the ground truth. mFDE$_k$ computes the minimum displacement error of the endpoint our predictions across the hypothesis with that of the ground truth. Miss$_{d,k}$ computes miss rate of top $k$ likely trajectories with the GT. A set of predictions is considered to be a miss if there's no hypothesis across predictions having maximum displacement point less than the threshold $d$. OffRoadRate computes percentage of output trajectories that fall outside the drivable region. We use the example API provided by Nuscenes to compute our metrics.
\vspace{-0.3cm}
\subsubsection{Quantitative Results}
We first show that ALAN can achieve on par or better performance compared to our baseline approaches. Here we evaluate ALAN with different anchor sampling strategies, top-M, oracle and best-of-all (BofA). In ALAN (top-M) we pick top $M$ trajectory outputs from different anchors based on predicted IOC scores for each trajectory. ALAN (oracle) uses oracle anchor with highest centerline score (see supplementary) and ALAN (BofA) picks best from top-k hypothesized lane anchors.  Results represented by Table \ref{tab:nu_bench} demonstrate that all our ALAN evaluations either show on par performance or significantly outperform other baselines on several metrics with at least 11\% improvements in terms of mADE$_{10}$ and 25\% boost in terms mFDE$_{10}$ from our BofA method. Moreover, all our ALAN predictions provide an OffRoadRate of $0.01$ showing only 1\% of the predicted trajectories fall outside the road. This is significantly lower compared to other baselines where they have 7\% or higher OffRoadRate's.  This strong coupling of output predictions with the semantics can be attributed to the anchor lanes that help in providing output predictions in the lane direction. Other approaches like \cite{multipath, covernet} use trajectories extracted from the train set, either as anchors or to perform classification, this can lead to poor generalization of outputs to unseen scenarios and trajectories with complex lane structure. Moreover, we would like to note that our ALAN performance is understated due issues such as unconnected lanes and places without lane centerlines in the data leading to bad anchors. We talk about such situations in supplementary but have not removed these here for benchmark purposes.

\vspace{-0.3cm}
\paragraph{Ablation Study:}
Further, we perform ablation studies of our ALAN along with the proposed DAC and other variants in Table \ref{tab:nu_ablation}. 
% As shown in Table \ref{tab:nu_ablation}, first we train a CVAE based single representational model similar to \cite{sriram2020smart} and compare it with a similar method trained using MCL\cite{stochasticMCL} objective. Our experiments show that MCL based methods are better at capturing diversity for a given number of samples as seen from their FDE values. Further, we introduce hypercolumn descriptors \cite{pixelnet} to extract multi-scale features and compare it with using a global context vector fed as input to the decoder.
We first introduce hypercolumn descriptors \cite{pixelnet} to extract multi-scale features and compare it with using a global context vector fed as input to the decoder.
% We see that using hypercolumn descriptors tend to give better performance in comparison to using global context vector as it captures nearby details from initial feature layers. 
Then we investigate several variants of our ALAN predictions. First, we add reference centerline as input and predict trajectories in $xy$ coordinate space (MCL + Poly). This improved the performance significantly. Using lane centerlines as anchors and predicting trajectories in $nt$ space (MCL+LA-$nt$) performed a little worse but we attribute this to networks difficulty in figuring out agent dynamics from anchor based inputs. For example, two trajectories with the same $nt$ coordinates can have different dynamics based on the lane that they're travelling. So we further add $xy$ coordinates as input and predict auxiliary trajectories in cartesian space (MCL+LA-$ntxy$). As it is shown in Table \ref{tab:nu_ablation}, making such auxiliary predictions improved the primary anchor based outputs. Further, we regularize our anchor outputs using auxiliary predictions and vice-versa. The intuition is that anchor outputs can benefit from auxiliary predictions when there's a bad input anchor since auxiliary predictions are not constrained to provide trajectories along the lane direction. Adding a regularizer to match our primary and auxiliary trajectories significantly improved our anchor output performance as seen in Table \ref{tab:nu_ablation} from MCL+LA-$ntxy$+Reg values.

 Comparing variations of ALAN in Table \ref{tab:nu_ablation}, it can be inferred that network trained with DAC beats the EWTA and RWTA objectives confirming the ability of the proposed DAC method to produce diverse hypotheses and capture the data distribution better. Please note that although we perform evaluations for DAC in trajectory prediction setting, MCL\cite{stochasticMCL} techniques are applicable in a wide range of problems where our DAC method can be used as a better initialization strategy for WTA objectives. 

% We then show comparisons of ALAN with our proposed DAC and EWTA objectives. From Table \ref{tab:nu_ablation} it can be inferred that network trained with DAC beats the baseline EWTA objective showing its ability to produce diverse hypothesis to capture the data distribution better. Please note that although we perform evaluations for DAC in trajectory prediction setting, MCL\cite{stochasticMCL} techniques are applicable in a wide range of problems where our DAC method can be used as a better initialization strategy for WTA objectives. 

% Further, use lane centerlines as anchors and predict trajectories in $nt$ space. The performance dropped a li

% The performance  slightly This can be attributed to network being unaware of the agent dynamics in  
% Then we add reference polyline as input to guide our network predictions. This already improved the performance significantly. In comparison, providing inputs with respect to the lane centerlines in $nt$ coordinates also improved 

% Notes while writing:\begin{itemize}
%     \item Mention DAC works in other problems not limited to TP. 
%     \item Multi-Agent predictions in one forward pass compared with other anchor methods.
%     \item OffRoad rate < 1 percent due to lane anchors and complex trajectories simplified due to nt coordinate prediction. 
% \end{itemize}
\vspace{-0.2cm}
\subsubsection{Qualitative Results}
\vspace{-0.1cm}
Figure \ref{fig:nu_img} shows qualitative results from ALAN. In general, using lane as anchors and transforming the prediction problem to $nt$ space can be helpful to guide the prediction and follow semantics. As we predict trajectories for a longer time horizon the executed trajectories become complex with more than just one straight or turn maneuvers where using lane as anchors can simplify the problem. 

% Figure \ref{fig:nu_img} shows qualitative results from ALAN. In general, using lane as anchors and transforming the prediction problem to $nt$ space can be helpful to guide the prediction and follow semantics. As we predict trajectories for a longer time horizon the executed trajectories become complex with more than just one straight or turn maneuvers where using lane as anchors can simplify the problem. 

% \begin{equation}
%     \texttt{Miss}_{d,k} = 
%     \begin{cases}
%     1, & if
%     \end{cases}
% \end{equation}

% Further, we evaluate our proposed DAC on Nuscenes prediction dataset. Here we investigate two network variants. First, a CVAE\cite{} architecture that takes in past trajectories and a top view map to predict output trajectories. Here instead of generally used variety loss \cite{Social_gan}. \sriram{Rewrite results for CVAE}

% The oracle error measures the diversity of our outputs predictions by choosing the closest hypothesis with the ground truth while the EMD distance quantifies the amount of probability mass that has to be moved from the predicted distribution to match the true distribution.

% \section{Outline}

% \begin{itemize}\itemsep -2pt
%     \item CVAE with polyline vs codes
%     \item MCL vs CVAE with polylines
%     \item Novel MCL initialization - Table 1\&2 Ablation for Divide and Conquer. CPI dataset and Nuscenes
%     \item Novel Coordinate Transformation - Ablation for Normal Tangential coordinate system
%     \item Nuscenes dataset benchmark.
%     \item Generalization - Argoverse Benchmark
% \end{itemize}

%% file: figures/cpi_img.tex
\begin{figure}
    \centering
    \begin{subfigure}{0.19\linewidth}
        \includegraphics[width=\linewidth]{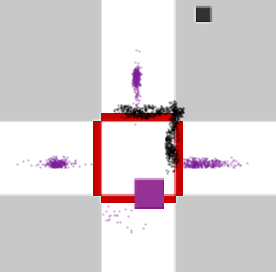}
        \caption{GT}
    \end{subfigure}
    \begin{subfigure}{0.19\linewidth}
        \includegraphics[width=\linewidth]{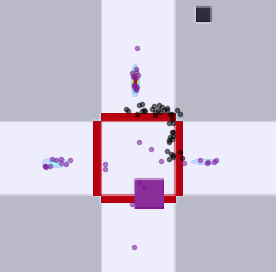}
        \caption{WTA}
    \end{subfigure}
    \begin{subfigure}{0.19\linewidth}
        \includegraphics[width=\linewidth]{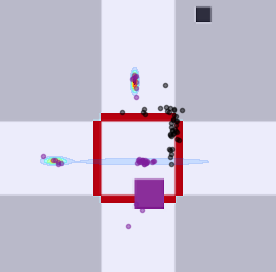}
        \caption{RWTA}
    \end{subfigure}
    \begin{subfigure}{0.19\linewidth}
        \includegraphics[width=\linewidth]{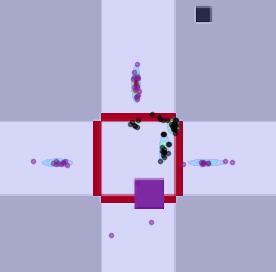}
        \caption{EWTA}
    \end{subfigure}
    \begin{subfigure}{0.19\linewidth}
        \includegraphics[width=\linewidth]{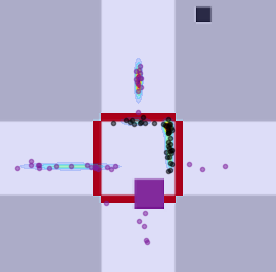}
        \caption{DAC}
    \end{subfigure}
    \caption{The figure illustrates predicted hypotheses and learned mixture distribution of goals using different WTA objectives on the CPI test set.  The purple and black box represent car and pedestrian at their current location. Predicted hypotheses are shown in their respective colours.
    % The purple box represents the car and black box represents the pedestrian at their current location. The predictions hypothesis are shown in their respective colours. 
    (e) captures the data distribution better with hypothesis spread out across the crosswalk resembling the ground truth distribution of points.}
    \label{fig:cpi_img}
\end{figure}

%% file: tables/cpi_table.tex
\begin{table}[]
\centering
\caption{\textbf{Comparion of Methods on CPI dataset based on FDE and EMD metrics, where p - pedestrian and c - car}}
\resizebox{\linewidth}{!}{
\begin{tabular}{ccccccc}
\toprule
Method & pFDE & cFDE & Avg FDE & pEMD & cEMD & Avg EMD \\ 
\midrule
DAC    & 5.56 & 5.61 & 5.58    & 1.14 & 1.48 & 1.31    \\
EWTA\cite{ewta}   & 5.8  & 5.63 & 5.76    & 1.09 & 1.59 & 1.34    \\
RWTA\cite{rwta}   & 4.90 & 9.56 & 7.23    & 1.02 & 1.64 & 1.33    \\
WTA\cite{stochasticMCL}    & 5.32 & 6.32 & 5.82    & 1.17 & 2.41 & 1.79    \\
CVAE   & 15.9 & 19.2 & 17.6    & 1.72 & 2.74 & 2.23    \\ 
\bottomrule
\end{tabular}
}
\label{tab:cpi_table}
\end{table}

%% file: figures/nu_img.tex
\begin{figure*}
    \centering
    \centering
    \begin{subfigure}{0.24\linewidth}
        \includegraphics[width=\linewidth]{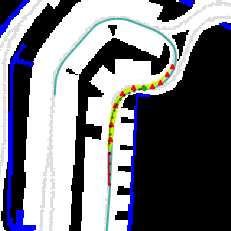}
        \caption{}
    \end{subfigure}
    \begin{subfigure}{0.24\linewidth}
        \includegraphics[width=\linewidth]{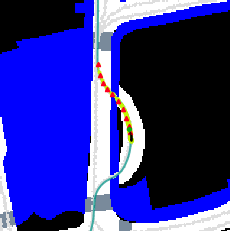}
        \caption{}
    \end{subfigure}
    \begin{subfigure}{0.24\linewidth}
        \includegraphics[width=\linewidth]{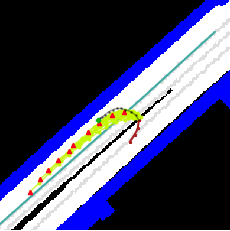}
        \caption{}
    \end{subfigure}
    \begin{subfigure}{0.24\linewidth}
        \includegraphics[width=\linewidth]{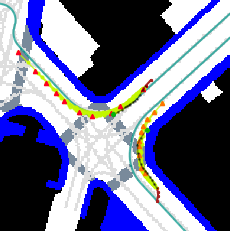}
        \caption{}
    \end{subfigure}
    \caption{Shows example predictions from ALAN. The past trajectory is shown in brown and the GT is shown in black. The endpoint of GT is shown as a green dot. The input lane anchor is shown in cyan, with predicted trajectories in green and their endpoints as triangles. (a) and (b) shows predictions that follow a complex lane structure. Anchor based predictions can be beneficial especially for a longer prediction horizon as the complexity of the trajectory increases anchors can be helpful in following the semantics. (c) predicts a U-turn with appropriate dynamics when the lane of interest is in opposite direction and (d) shows a multi-agent prediction scenario.}
    \label{fig:nu_img}
\end{figure*}

%% file: tables/nuscenes_benchmark.tex
\begin{table*}[]
\centering
\caption{\textbf{Nuscenes Trajectory Prediction Benchmark}}
\resizebox{\textwidth}{!}{
\begin{tabular}{cccccccccc}
\toprule
Model                & mADE\_1 & mADE\_5 & mADE\_10 & Miss\_2\_5 & Miss\_2\_10 & mFDE\_1 & mFDE\_5 & mFDE\_10 & OffRoadRate \\
\midrule
cxx                     & -    & {\bf 1.63} & 1.29 & 69 & 60 & \underline{8.86} & -    & -    & 0.08 \\
pq                      & -    & 2.23 & 1.68 & 69 & 56 & 9.56 & -    & -    & 0.12 \\
CoverNet\cite{covernet}                & -    & 2.62 & 1.92 & 76 & 64 & 11.36 & -    & -    & 0.13 \\
MTP \cite{mtp}                     & \underline{4.42} & 2.22 & 1.74 & 74 & 67 & 10.36 & 4.83 & 3.54 & 0.25 \\
MultiPath \cite{multipath}               & 4.43 & 1.78 & 1.55 & 78 & 76 & 10.16 & 3.62 & 2.93 & 0.36 \\ 
Trajectron++\cite{trajectronpp}            & -    & 1.88 & 1.51 & 70 & 57 & 9.52  & -    & -    & 0.25 \\
MHA\_JAM \cite{mhajam}              & {\bf 3.69} & 1.81 & 1.24 & \underline{59} & \underline{45} & {\bf 8.57}  & 3.72 & 2.21 & \underline{0.07} \\
\midrule
\textbf{ALAN} (top-M)          & 4.62 & 1.87 & 1.22 & 60 & 49 & 9.98 & 3.54 & 1.87 & {\bf 0.01}\\
\textbf{ALAN} (Oracle)        & 4.61 & 1.78 & \underline{1.16} & 59 & 48 & 9.95 & {\bf 3.29} & \underline{1.70} & {\bf 0.01}\\
\textbf{ALAN} (BofA)          & 4.67 & \underline{1.77} & {\bf 1.10} & {\bf 57} & {\bf 45} & 10.0 & \underline{3.32} &	{\bf 1.66} & {\bf 0.01}\\
% \midrule
% \textbf{ALAN}    & 6.91	& 1.96  & 1.22 & 77 & 52 & 14.41 & 3.68 & 1.86 & 0.01\\
% \textbf{ALAN (Oracle)}     & 6.94	& 1.87	& 1.16	& 75 & 49 & 14.4 & 3.46 & 1.7 & 0.01\\ 
% \midrule
% \textbf{ALAN (Oracle)} & 4.31 & 2.10 & 1.17 & 63 & 50 & 9.06  & 3.98 & 1.73 & 0.01 \\
% \textbf{ALAN (BofA)} & 4.28 & 2.04 & 1.11 & 61 & 46 & 9.02  & 3.92 & 1.66 & 0.01 \\
\bottomrule
\end{tabular}
}
\label{tab:nu_bench}
\end{table*}

%% file: tables/nuscenes_ablation.tex
\begin{table*}[]
\centering
\caption{\textbf{Ablation Study on Nuscenes dataset}}
\resizebox{\textwidth}{!}{
\begin{tabular}{cccccccccc}
\toprule
Model                & mADE\_1 & mADE\_5 & mADE\_10 & Miss\_2\_5 & Miss\_2\_10 & mFDE\_1 & mFDE\_5 & mFDE\_10 & OffRoadRate \\ 
\midrule
% CVAE + Codes \textbf{(SMART)} & & & & & & & & & \\
% CVAE + PL & & & & & & & & & \\
% MCL + PL & & & & & & & & & \\
% MCL + PL + EWTA & & & & & & & & & \\
% MCL + PL + DAC & & & & & & & & & \\
% MCL + PL + LA + DAC \textbf{(SMART++)} & & & & & & & & & \\ 
% CVAE - WTA                  & 5.93 & 2.26 & 1.69 & 77 & 64 & 13.03 & 4.57 & 3.10 & 0.04\\
% CVAE - EWTA                 & 5.55 & 2.20 & 1.65 & 76 & 62 & 12.32 & 4.53 & 3.10 & 0.03\\
CVAE                  & 5.51 & 2.12 & 1.55 & 76 & 62 & 12.03 & 4.45 & 2.85 & 0.03\\
MCL + Global          & 8.45 & 2.85	& 1.88 & 87 & 75 & 17.52 & 5.34 & 3.05 & 0.16\\ 
MCL + Hyper           & 5.55 & \textbf{1.99}	& 1.33 & 72 & 58 & 12.11 & \textbf{3.81}	& 2.26 & 0.12\\
% MCL + Poly                  & 6.96 & 2.26 & 1.39 & 82 & 65 & 14.46 & 4.25 & 2.21 & 0.06\\
MCL + Poly                  & 6.50 & 2.03 & 1.27 & 77 & 57 & 13.6 & 3.88 & 2.01 & 0.05\\
MCL + LA - $nt$      & 4.69 & 2.62 & 1.45 & 78 & 59 & 9.86  & 4.83 & 2.26 & 0.05\\
% MCL + Poly - EWTA           & 6.50 & 2.03 & 1.27 & 77 & 57 & 13.60 & 3.88 & 2.03 & 0.05\\
% MCL + Poly - DAC            & 6.48 & 1.96 & 1.25 & 76 & 52 & 13.70 & 3.71 & 2.00 & 0.04\\
% MCL + Poly + LA - $nt$\\
MCL + LA - $ntxy$    & 6.65 & 2.14 & 1.41 & 75 & 53 & 13.86 & 3.97 & 2.18 & 0.01\\
MCL + LA - $ntxy$ + Reg. + WTA & 7.45 & 3.91 & 1.72 & 82 & 71 & 13.5 & 6.49 & 2.37 & 0.01\\
MCL + LA - $ntxy$ + Reg. + RWTA & 4.41 & 2.55 & 1.21 & 64 & \textbf{45} & 9.22 & 5.03 & 1.77 & 0.01\\
MCL + LA - $ntxy$ + Reg. + EWTA & 4.38 & 2.06 & 1.20 & 64 & 52 & 9.16 & 3.83 & 1.76 & 0.01\\
MCL + LA - $ntxy$ + Reg. + DAC  & \textbf{4.31} & 2.10 & \textbf{1.17} & \textbf{63} & 50 & \textbf{9.06} & 3.98 & \textbf{1.73} & \textbf{0.01}\\
\bottomrule
\end{tabular}
}
\label{tab:nu_ablation}
\end{table*}

%% file: sec_conclusion.tex
\vspace{-0.03cm}
\section{Conclusion}

In this paper we addressed issues related to learning multi-modal outputs using WTA objectives and using driving knowledge to impose constraints on output predictions. First, we introduced a novel DAC approach that learns diverse hypotheses to capture the data distribution without any spurious modes. Further, we introduced ALAN that provides diverse and context aware trajectories using anchor lanes. 
Our experiments on both synthetic and real data demonstrated the superiority of our proposed DAC method in learning multi-modal outputs.
In addition, we demonstrated that using lane anchors can be helpful in providing accurate predictions with strong semantic coupling.

% In this paper we addressed issues related to learning multi-modal outputs using WTA objectives and imposing constraints on output predictions from driving knowledge. First, we introduced a novel Divide-And-Conquer approach that learns diverse hypotheses to capture the data distribution without any spurious modes. 
% Further, we introduced ALAN framework that provides diverse, context aware, multi-agent trajectories using anchor lanes. 
% Our experiments on both synthetic and real data demonstrated the superiority of our proposed DAC method in learning multi-modal outputs.
% In addition, we demonstrated that using lane anchors can be helpful in providing accurate predictions with strong semantic coupling.

% In this paper we have addressed issues related to learning multi-modal outputs using winner-takes-all objective and imposing constraints on prediction outputs based on driving knowledge. First, we introduced a novel Divide-And-Conquer approach that learns diverse hypotheses to capture the data distribution without any spurious modes. 
% Further, we introduced ALAN framework that provides diverse, context aware, multi-agent trajectories using lane anchors. 
% Our experiments on both synthetic and real data demonstrated the superiority of our proposed DAC method in learning multi-modal outputs.
% In addition, we demonstrated that using lane anchors can be helpful in providing accurate predictions with strong semantic coupling.

%% file: sec_supp_arch.tex
\section{Further Details for ALAN}

\subsection{Network Architecture}

In this section, we provide more details about our network architecture. 
\vspace{-0.3cm}
\paragraph{Centerline Encoder:} It takes lane anchor for each individual agent in the form of $p\times2$ points and reshapes them as $2p\times1\times1$ vector. We then pass it through a series of $1\times1$ convolutions with 256,128,64 output filters. 
% \paragraph{Centerline Encoder:} It takes in the lane anchor in the form of $p\times2$ points and reshapes as $2p\times1\times1$ vector. We then pass it through a series of $1\times1$ convolutions with 256,128,64 output filters. 

\vspace{-0.3cm}
\paragraph{Past Trajectory Encoder:} It contains an embedding layer and an LSTM encoder. The embedding layer takes a 5 dimensional vector containing $X_i^t, N_{i,k}^t$ for every timestep along with a boolean mask indicating if trajectory information is available for the timestep and produces an embedded vector of 16 dimensions. Then the embedded past trajectory is fed through a LSTM of hidden size 64 to produce a past state vector of 64 dimensions.

\vspace{-0.3cm}
\paragraph{Multi-Agent Convolutional Encoder:} The past state vector (64 dim) and embedded centerline vector (64 dim) are concatenated to form a 128 dimensional vector for every agent. The agent specific information is then  encoded at its respective location to form a $H\times W\times 128$ vector. This along with the BEV map of size $H\times W \times 3$ is provided as input to this module. The module uses a ResNet-18\cite{resnet18} encoder backbone where we replace the first layer with a convolutions of $3\times3$ kernel size and stride 2. We then pass the features through ResNet\cite{resnet18} base layers from 1 to 7. 

\vspace{-0.3cm}
\paragraph{Hypercolumn Trajectory Decoder:} First we extract hypercolumn descriptors for every agent based on the agent's location in the BEV map. Specifically, we extract hypercolumn features from layers 0,2,4,5,6,7 and pass them through a series of $1\times1$ convolutions containing 2048,1024,1024 filters. Finally, a $1\times1$ output convolution then produces primary and auxiliary outputs.

\vspace{-0.3cm}
\paragraph{Ranking Module:} It takes in 1024 features vectors from the Hypercolumn Trajectory Decoder before the final output layer and passes them through a $1\times1$ convolution with 1024 filters and finally produces $M$ trajectory scores as output. 

\subsection{Learning}

Our input BEV map is of size $256 \times 256$ dimensions at a resolution of $0.5m$ per pixel. The models are trained with Adam\cite{adam} optimizer with an initial learning rate of 1e-4 and batch size 8. We use exponential lr decay with gamma value 0.95 called after every epoch. The hypotheses are split in case of DAC after every 2000 iterations and the models are trained for 150k iterations (approximately 38 epochs). The ALAN is implemented in pytorch\cite{pytorch} and trained on NVIDIA RTX 2080Ti GPU. 
% We use a BEV map of size $256 \times 256$ at a resolution of $0.5m$ per pixel. The models are trained with Adam\cite{adam} optimizer with a initial learning rate of 1e-4 and batch size 8. We use an exponential lr decay with gamma value 0.95 called after every epoch. We split hypothesis for DAC after every 2000 iterations. We train the models for 150k iterations approximately 38 epochs. We implement ALAN in pytorch\cite{pytorch} and train on NVIDIA RTX 2080Ti GPU. 

\subsection{Nuscenes Dataset}
Approximately 2.5\% of validation set contains bad anchors, such as some due to either unconnected lanes or places without lane centerlines. In such cases, for the benchmark evaluation we use the nearest lane centerline that is closest to the trajectory. Further, we evaluate our method by approximately removing examples which have average normal distance of the past trajectory to nearest lane greater than a threshold distance (3 meters) and compare it with baselines \cite{covernet, mtp}. As seen from Table \ref{tab:nu_supp_bench} our proposed ALAN provides even better performance when bad anchors are removed. 
% As seen from Table \ref{tab:nu_supp_bench} our proposed ALAN provides better performance numbers in general with bad anchors removed and also in comparison to baselines \cite{covernet, mtp}.
\input{./tables/nuscenes_supp_bench}

%% file: tables/nuscenes_supp_bench.tex
\begin{table*}[]
\centering
\caption{Nuscenes Evaluations. Removing bad anchors based on normal distance to lane (threshold = 3m). Total validation examples after removing = 8823.}
\resizebox{\textwidth}{!}{
\begin{tabular}{cccccccccc}
\toprule
Model                & mADE\_1 & mADE\_5 & mADE\_10 & Miss\_2\_5 & Miss\_2\_10 & mFDE\_1 & mFDE\_5 & mFDE\_10 & OffRoadRate \\
\midrule

CoverNet\cite{covernet}                & 6.81 & 3.09 & 2.39 & 87 & 76 & 13.9 & 5.92 & 4.27 & 0.13\\
% CoverNet\cite{covernet}                & -    & 2.62 & 1.92 & 76 & 64 & 11.36 & -    & -    & 0.13 \\
MTP \cite{mtp}                     & 4.14 & 2.80 & 1.74 & 70 & 53 & 9.91 & 6.42 & 3.71 & 0.11\\
% MTP \cite{mtp}                     & 4.42 & 2.22 & 1.74 & 74 & 67 & 10.36 & 4.83 & 3.54 & 0.25 \\
\textbf{ALAN} (top-M)          & 4.49 & 1.79 & 1.14 & 58 & 47 & 9.76 & 3.41 & 1.76 & 0.003\\
\textbf{ALAN} (Oracle)        & 4.48 & 1.70 & 1.08 & 57 & 46 & 9.72 & 3.16 & 1.60 & 0.004\\
\textbf{ALAN} (BofA)          & 4.54 & 1.69 & 1.03 & 55 & 43 & 9.84 & 3.20 & 1.56 & 0.006\\
% \midrule
% \textbf{ALAN}    & 6.91	& 1.96  & 1.22 & 77 & 52 & 14.41 & 3.68 & 1.86 & 0.01\\
% \textbf{ALAN (Oracle)}     & 6.94	& 1.87	& 1.16	& 75 & 49 & 14.4 & 3.46 & 1.7 & 0.01\\ 
% \midrule
% \textbf{ALAN (Oracle)} & 4.31 & 2.10 & 1.17 & 63 & 50 & 9.06  & 3.98 & 1.73 & 0.01 \\
% \textbf{ALAN (BofA)} & 4.28 & 2.04 & 1.11 & 61 & 46 & 9.02  & 3.92 & 1.66 & 0.01 \\
\bottomrule
\end{tabular}
}
\label{tab:nu_supp_bench}
\end{table*}

%% file: sec_supp_poly.tex
\section{Anchor Retrieval}

Nuscenes\cite{nuscenes2019} provides HD map data containing lane centerline information represented a sequence of points. We follow steps similar to \cite{whatif} in order to retrieve plausible lanes. 

\textbf{Identify Closest Lanes} Given a position of the vehicle in city coordinates we first identify a set of closest lane segments to the vehicle within a radius $d$.

\textbf{Retrieve Candidate Anchors} We identify plausible lane anchors by traversing through successor and predecessor lanes till a threshold distance. Several connected lane segments from an anchor.

\textbf{Prune Duplicate Anchors} We then filter candidate anchors by removing lanes, which are a subset of others.

\textbf{Heuristic based Pruning} Based on the vehicle's velocity we identify a look ahead point on the lane for a future timestep $T$ and remove duplicate anchors which pass through the same point. This is done to further reduce the number of plausible anchors and remove duplicates which only diverge after a sufficient distance from the vehicle's position. 

\textbf{Distance Along Lane Score} Then we rank the candidates anchors based on distance travelled along the lane by the vehicle. First, we calculate the corresponding $nt$ coordinates for the trajectory along every plausible anchor. The score is determined as the absolute sum of normal values for every timestep in the trajectory. Anchors are then ranked based on their scores. 

\textbf{Centerline Yaw Score} Further, we rank candidate anchors which have same distance along lane score based on centerline yaw score. It is calculated as the absolute difference between yaw angle of the vehicle and lane yaw angle at a point closest to the vehicle.

\textbf{Learning} Every plausible anchor divided into $p$ equally spaced points, in our case $p=150$. For training, we use the oracle anchor where oracle is determined based on the trajectory information from $1...T$. During inference, for ALAN (top-M) we rank trajectories using observed locations from $1...t_{obs}$.

%% file: sec_supp_qualitative.tex
\section{Additional Results}

In this section, we show some additional quantitative and qualitative comparisons for the proposed DAC and ALAN framework. 

\subsection{DAC Qualitative Comparisons}

We evaluate our proposed DAC through additional simulations including modes with non-uniform probabilities and compare it with previous WTA objectives \cite{stochasticMCL, rwta, ewta} (shown in Figure \ref{fig:toy2_dac}). 
% As shown, Figure \ref{fig:toy2_dac} contains mixture distribution with non-uniform likelihoods. 
As observed, WTA\cite{stochasticMCL} leaves many hypothesis untrained. While RWTA\cite{rwta} solves the convergence problem in WTA it brings non-winner hypothesis to an equilibrium position due to residual constraints. Further, EWTA\cite{ewta} captures the data distribution better but still suffers from the problem of spurious modes as hypothesis can be attracted towards many ground truths and finally left untrained as only top k hypotheses are penalized. Finally, proposed DAC captures the data distribution as good or better even when modes have such low likelihoods and solves the problem of spurious modes by making use of all hypotheses. In every stage, DAC reaches close to a Centroidal Voronoi Tessellation\cite{centroidalVoronoi} with effective number of outputs increasing at every stage, leading to hypotheses capturing some probability mass and thus avoiding the spurious mode problem. 

% Further, we evaluate our proposed DAC through additional simulations including modes with non-uniform probability and compare it with previous WTA objectives \cite{stochasticMCL, rwta, ewta}. As seen Figure \ref{fig:toy2_dac} WTA\cite{stochasticMCL} leaves many hypothesis untrained. While RWTA\cite{rwta} solves the convergence problem in WTA it brings non-winner hypothesis to an equilibrium position due to residual constraints. Further, EWTA\cite{ewta} captures the data distribution better but still suffers from the problem of spurious modes as hypothesis can be attracted towards many ground truths and finally left untrained as only top k hypotheses are penalized. Finally, proposed DAC captures the data distribution better and solves the problem of spurious modes by making use of all hypotheses. As in every stage, DAC reaches close to a Centroidal Voronoi Tessellation\cite{centroidalVoronoi} with effective number of outputs increasing at every stage, leading to hypotheses capturing some probability mass and thus avoiding the spurious mode problem. 

% In Figure \ref{fig:toy2_dac} we also evaluate 

% \sriram{Editing... Adding information about non-uniform modes. To be added.. DAC on other datasets (DeeplabV3 results)}

\subsection{DAC on Other Networks}

Further, we implement our proposed DAC on other popular networks such as DeepLabV3\cite{deeplab} where we train and test on PascalVOC2012\cite{pascal-voc-2012} dataset for semantic segmentation task and calculate the IoU scores for 21 classes. Our mIoU scores across all clases is reported in Table \ref{tab:deeplab}.  Here, we modify the final layer of DeepLabV3\cite{deeplab} to contain multiple segmentation heads in order to implement DAC and WTA. 

\input{./tables/supp_seg_table}

\input{./figures/dac_toy2}

\subsection{ALAN Qualitative Comparisons}

Figure \ref{fig:nu_supp_img} shows qualitative comparison of ALAN with other baselines \cite{covernet, mtp}.
% Figure \ref{fig:nu_supp_img} shows ALAN prediction in comparison to other baselines\cite{covernet,mtp}. As observed, ALAN predictions are much more semantically aligned and accurate in comparison to the baselines. This strong semantic coupling can be attributed to using lane centerlines as anchors and transforming the problem to $nt$ space along the direction of lanes. 
\input{./figures/supp_nu_img}

%% file: tables/supp_seg_table.tex
\begin{table}[!!t]
\resizebox{\linewidth}{!}{
\begin{tabular}{|c|c|c|c|}
\hline
Method (mIoU)                       & DeepLabV3 & DeepLabV3 + WTA & DeepLabV3 + DAC \\ \hline
Pascal2012 (val set)                & 76.83     & 82.14           & 83.44           \\ \hline
\end{tabular}
}
\caption{mIoU score for DeepLabV3\cite{deeplab} on Pascal2012\cite{pascal-voc-2012} dataset with no multimodality, WTA(k=3) and DAC on Pascal\cite{pascal-voc-2012} val set.}
\label{tab:deeplab}
\end{table}

%% file: figures/dac_toy2.tex
\begin{figure*}[t]
    \centering
    \begin{subfigure}{0.24\linewidth}
        \includegraphics[width=\linewidth]{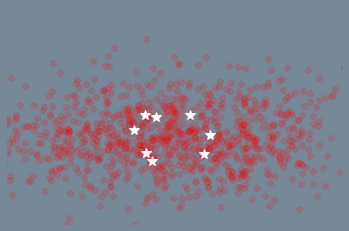}
        \caption{Init}
        % \label{fig:toy_dac_init}
    \end{subfigure}
    \begin{subfigure}{0.24\linewidth}
        \includegraphics[width=\linewidth]{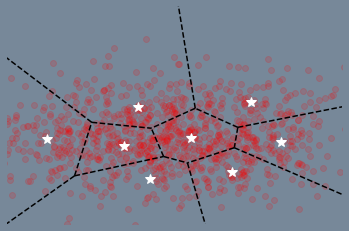}
        \caption{WTA}
        % \label{fig:toy_dac_wta}
    \end{subfigure}
    \begin{subfigure}{0.24\linewidth}
        \includegraphics[width=\linewidth]{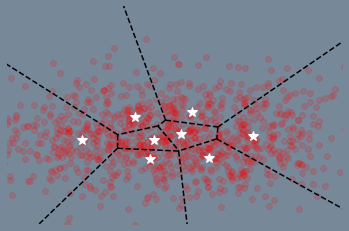}
        \caption{RWTA}
        % \label{fig:toy_dac_rwta}
    \end{subfigure}
    \begin{subfigure}{0.24\linewidth}
        \includegraphics[width=\linewidth]{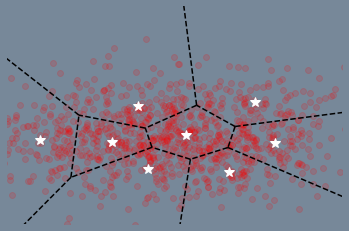}
        \caption{EWTA}
        % \label{fig:toy_dac_ewta}
    \end{subfigure}

    \begin{subfigure}{0.24\linewidth}
        \includegraphics[width=\linewidth]{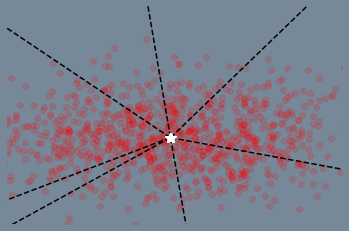}
        \caption{DAC - Depth 1}
        % \label{fig:toy_dac_d1}
    \end{subfigure}
    \begin{subfigure}{0.24\linewidth}
        \includegraphics[width=\linewidth]{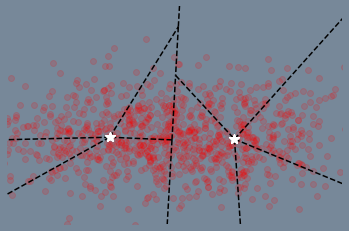}
        \caption{DAC - Depth 2}
        % \label{fig:toy_dac_d2}
    \end{subfigure}
    \begin{subfigure}{0.24\linewidth}
        \includegraphics[width=\linewidth]{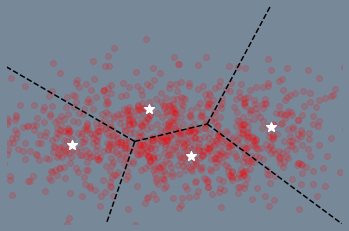}
        \caption{DAC - Depth 3}
        % \label{fig:toy_dac_d3}
    \end{subfigure}
    \begin{subfigure}{0.24\linewidth}
        \includegraphics[width=\linewidth]{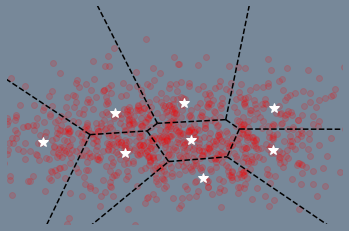}
        \caption{DAC - Depth 4}
        % \label{fig:toy_dac_d4}
    \end{subfigure}
    
    \begin{subfigure}{0.24\linewidth}
        \includegraphics[width=\linewidth]{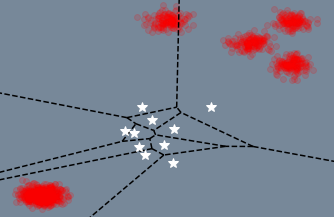}
        \caption{Init}
        % \label{fig:toy_dac_init}
    \end{subfigure}
    \begin{subfigure}{0.24\linewidth}
        \includegraphics[width=\linewidth]{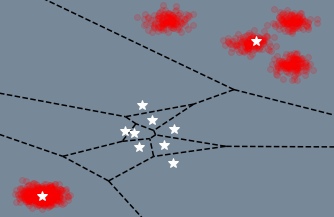}
        \caption{WTA}
        % \label{fig:toy_dac_wta}
    \end{subfigure}
    \begin{subfigure}{0.24\linewidth}
        \includegraphics[width=\linewidth]{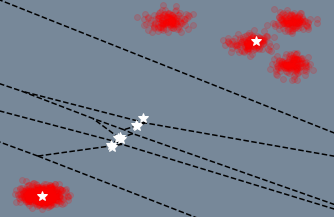}
        \caption{RWTA}
        % \label{fig:toy_dac_rwta}
    \end{subfigure}
    \begin{subfigure}{0.24\linewidth}
        \includegraphics[width=\linewidth]{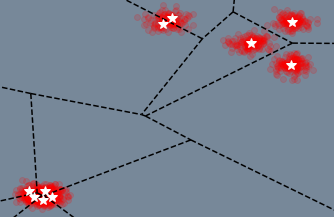}
        \caption{EWTA}
        % \label{fig:toy_dac_ewta}
    \end{subfigure}

    \begin{subfigure}{0.24\linewidth}
        \includegraphics[width=\linewidth]{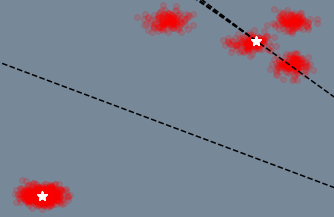}
        \caption{DAC - Depth 2}
        % \label{fig:toy_dac_wta}
    \end{subfigure}
    \begin{subfigure}{0.24\linewidth}
        \includegraphics[width=\linewidth]{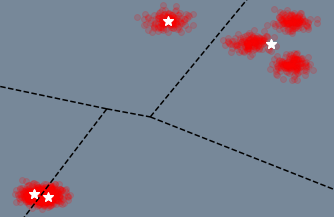}
        \caption{DAC - Depth 3}
        % \label{fig:toy_dac_rwta}
    \end{subfigure}
    \begin{subfigure}{0.24\linewidth}
        \includegraphics[width=\linewidth]{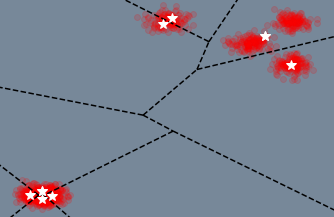}
        \caption{DAC - Depth 4}
        % \label{fig:toy_dac_ewta}
    \end{subfigure}
    \begin{subfigure}{0.24\linewidth}
        \includegraphics[width=\linewidth]{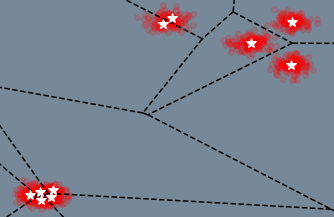}
        \caption{DAC - Depth 5}
        % \label{fig:toy_dac_init}
    \end{subfigure}
    
    \begin{subfigure}{0.24\linewidth}
        \includegraphics[width=\linewidth]{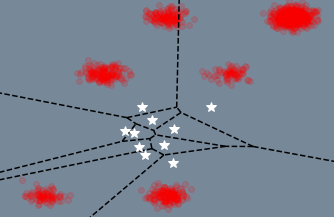}
        \caption{Init}
        % \label{fig:toy_dac_init}
    \end{subfigure}
    \begin{subfigure}{0.24\linewidth}
        \includegraphics[width=\linewidth]{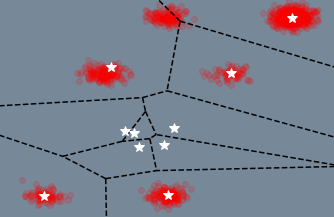}
        \caption{WTA}
        % \label{fig:toy_dac_wta}
    \end{subfigure}
    \begin{subfigure}{0.24\linewidth}
        \includegraphics[width=\linewidth]{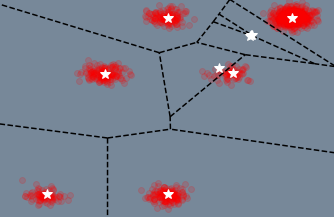}
        \caption{RWTA}
        % \label{fig:toy_dac_rwta}
    \end{subfigure}
    \begin{subfigure}{0.24\linewidth}
        \includegraphics[width=\linewidth]{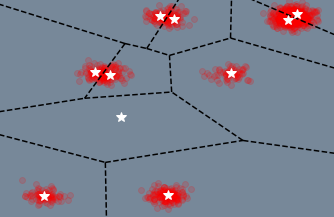}
        \caption{EWTA}
        % \label{fig:toy_dac_ewta}
    \end{subfigure}

    \begin{subfigure}{0.24\linewidth}
        \includegraphics[width=\linewidth]{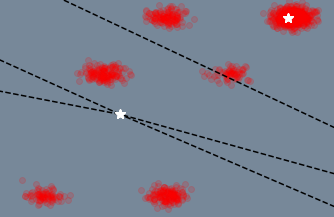}
        \caption{DAC - Depth 2}
        % \label{fig:toy_dac_wta}
    \end{subfigure}
    \begin{subfigure}{0.24\linewidth}
        \includegraphics[width=\linewidth]{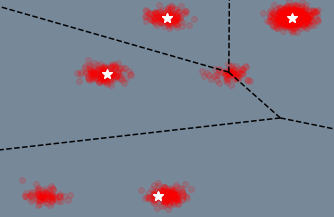}
        \caption{DAC - Depth 3}
        % \label{fig:toy_dac_rwta}
    \end{subfigure}
    \begin{subfigure}{0.24\linewidth}
        \includegraphics[width=\linewidth]{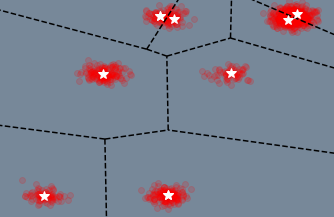}
        \caption{DAC - Depth 4}
        % \label{fig:toy_dac_ewta}
    \end{subfigure}
    \begin{subfigure}{0.24\linewidth}
        \includegraphics[width=\linewidth]{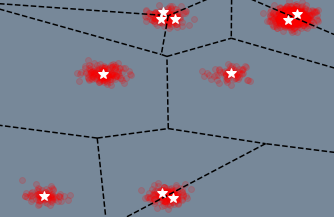}
        \caption{DAC - Depth 5}
        % \label{fig:toy_dac_init}
    \end{subfigure}
    
    \caption{Additional toy examples including modes with non-uniform likelihoods comparing DAC with other WTA family of objectives. The first two rows shows a mixture distribution with high variance and second two rows show a well separated gaussian mixture model. Other examples contain modes with $\pi = \{5,7.5,12.5,50\}\%$. As seen from Row 3, WTA captures the distribution poorly by either leaving many hypotheses untrained or by introducing spurious modes that do not correspond to the data distribution. On the other hand, every hypothesis in proposed DAC captures some part of the data as seen from its voronoi space.
    % On the other hand, proposed DAC captures the distribution better by splitting the hypothesis and forming a centroidal voronoi tessellation at every stage. Every hypothesis in DAC captures some part of the data as seen from its voronoi space.
    }
    \label{fig:toy2_dac}
\end{figure*}

%% file: figures/supp_nu_img.tex
\begin{figure*}
    \centering
    \begin{subfigure}{0.24\linewidth}
        \includegraphics[width=\linewidth, trim={{0.5\linewidth} {0.5\linewidth} {0.5\linewidth} {0.5\linewidth}}, clip]{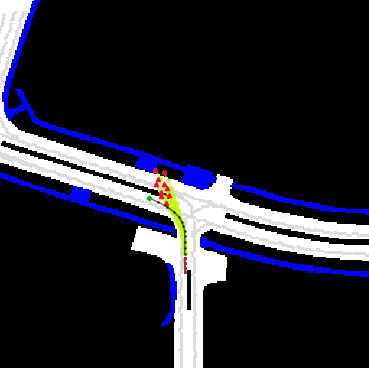}
        % \caption{}
    \end{subfigure}
    \begin{subfigure}{0.24\linewidth}
        \includegraphics[width=\linewidth, trim={{0.5\linewidth} {0.5\linewidth} {0.5\linewidth} {0.5\linewidth}}, clip]{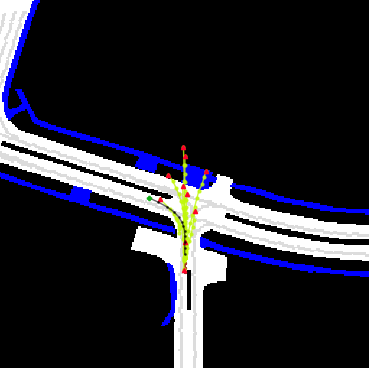}
        % \caption{}
    \end{subfigure}
    \begin{subfigure}{0.24\linewidth}
        \includegraphics[width=\linewidth, trim={{0.5\linewidth} {0.5\linewidth} {0.5\linewidth} {0.5\linewidth}}, clip]{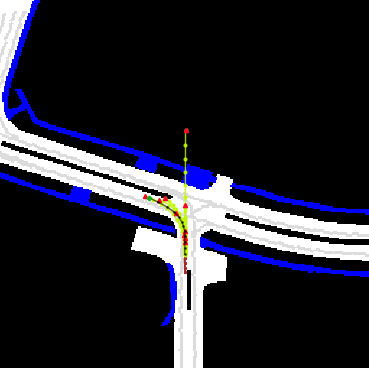}
        % \caption{}
    \end{subfigure}
    \begin{subfigure}{0.24\linewidth}
        \includegraphics[width=\linewidth, trim={{0.7\linewidth} {0.7\linewidth} {0.7\linewidth} {0.7\linewidth}}, clip]{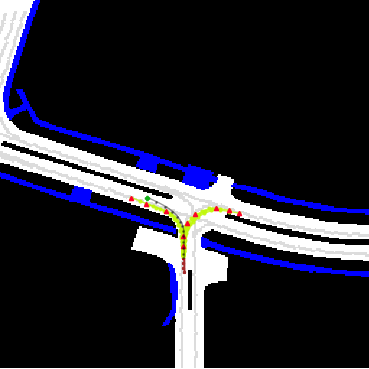}
        % \caption{}
    \end{subfigure}
    
    \begin{subfigure}{0.24\linewidth}
        \includegraphics[width=\linewidth, trim={{0.5\linewidth} {0.5\linewidth} {0.5\linewidth} {0.5\linewidth}}, clip]{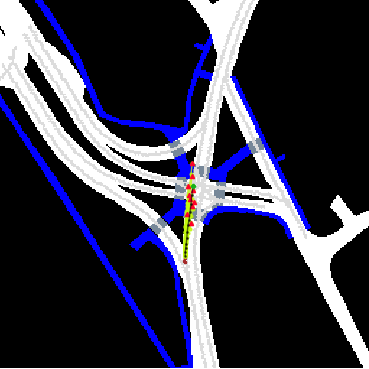}
        % \caption{}
    \end{subfigure}
    \begin{subfigure}{0.24\linewidth}
        \includegraphics[width=\linewidth, trim={{0.5\linewidth} {0.5\linewidth} {0.5\linewidth} {0.5\linewidth}}, clip]{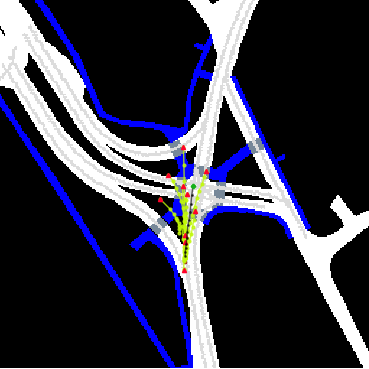}
        % \caption{}
    \end{subfigure}
    \begin{subfigure}{0.24\linewidth}
        \includegraphics[width=\linewidth, trim={{0.5\linewidth} {0.5\linewidth} {0.5\linewidth} {0.5\linewidth}}, clip]{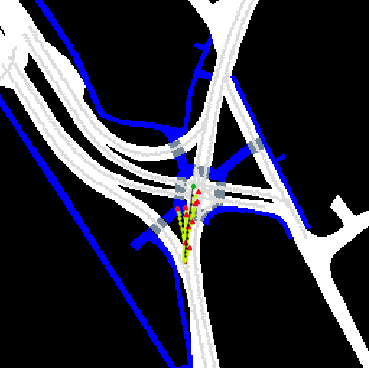}
        % \caption{}
    \end{subfigure}
    \begin{subfigure}{0.24\linewidth}
        \includegraphics[width=\linewidth, trim={{0.7\linewidth} {0.7\linewidth} {0.7\linewidth} {0.7\linewidth}}, clip]{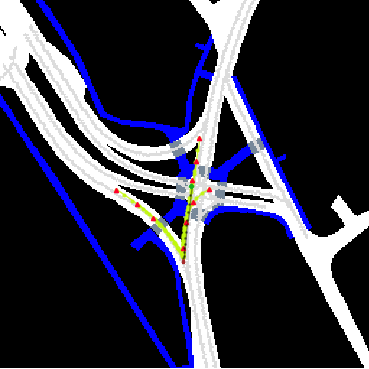}
        % \caption{}
    \end{subfigure}
    
    \begin{subfigure}{0.24\linewidth}
        \includegraphics[width=\linewidth, trim={{0.5\linewidth} {0.5\linewidth} {0.5\linewidth} {0.5\linewidth}}, clip]{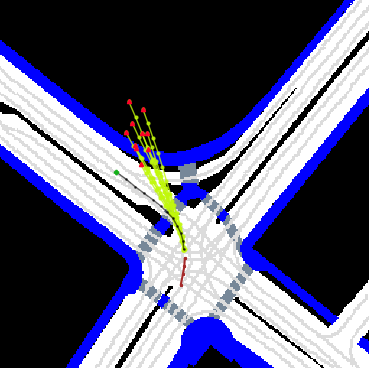}
        % \caption{}
    \end{subfigure}
    \begin{subfigure}{0.24\linewidth}
        \includegraphics[width=\linewidth, trim={{0.5\linewidth} {0.5\linewidth} {0.5\linewidth} {0.5\linewidth}}, clip]{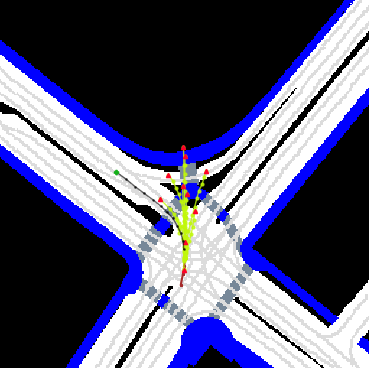}
        % \caption{}
    \end{subfigure}
    \begin{subfigure}{0.24\linewidth}
        \includegraphics[width=\linewidth, trim={{0.5\linewidth} {0.5\linewidth} {0.5\linewidth} {0.5\linewidth}}, clip]{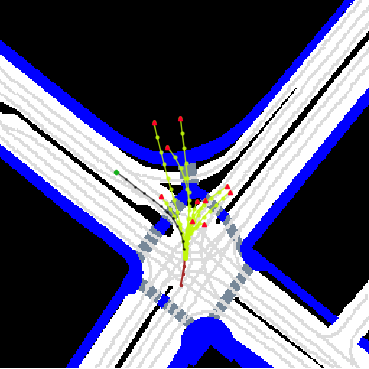}
        % \caption{}
    \end{subfigure}
    \begin{subfigure}{0.24\linewidth}
        \includegraphics[width=\linewidth, trim={{0.7\linewidth} {0.7\linewidth} {0.7\linewidth} {0.7\linewidth}}, clip]{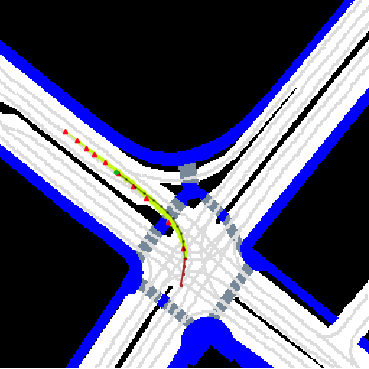}
        % \caption{}
    \end{subfigure}
    
    \begin{subfigure}{0.24\linewidth}
        \includegraphics[width=\linewidth, trim={{0.5\linewidth} {0.5\linewidth} {0.5\linewidth} {0.5\linewidth}}, clip]{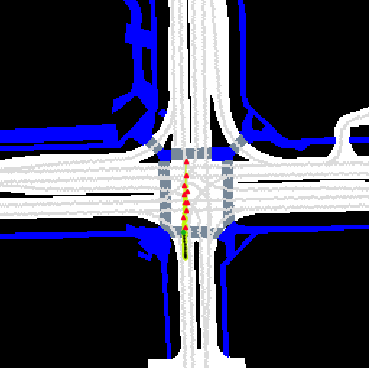}
        % \caption{}
    \end{subfigure}
    \begin{subfigure}{0.24\linewidth}
        \includegraphics[width=\linewidth, trim={{0.5\linewidth} {0.5\linewidth} {0.5\linewidth} {0.5\linewidth}}, clip]{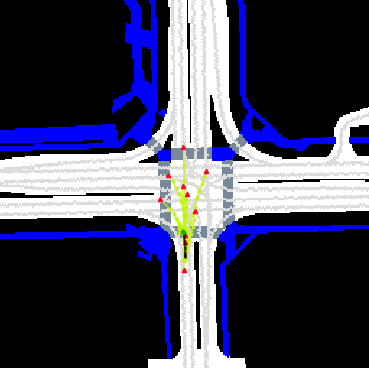}
        % \caption{}
    \end{subfigure}
    \begin{subfigure}{0.24\linewidth}
        \includegraphics[width=\linewidth, trim={{0.5\linewidth} {0.5\linewidth} {0.5\linewidth} {0.5\linewidth}}, clip]{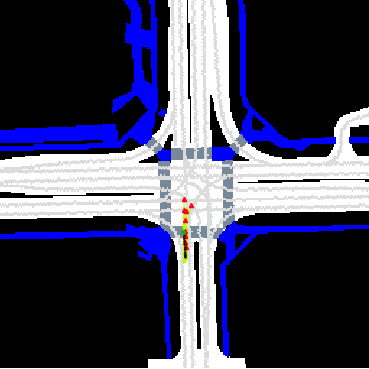}
        % \caption{}
    \end{subfigure}
    \begin{subfigure}{0.24\linewidth}
        \includegraphics[width=\linewidth, trim={{0.7\linewidth} {0.7\linewidth} {0.7\linewidth} {0.7\linewidth}}, clip]{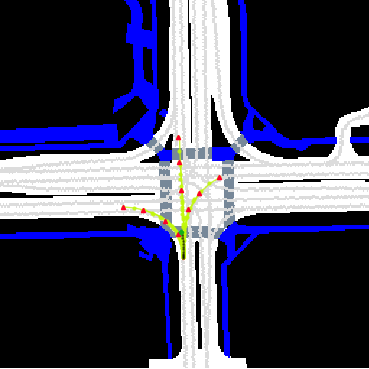}
        % \caption{}
    \end{subfigure}
    
    \begin{subfigure}{0.24\linewidth}
        \includegraphics[width=\linewidth, trim={{0.5\linewidth} {0.5\linewidth} {0.5\linewidth} {0.5\linewidth}}, clip]{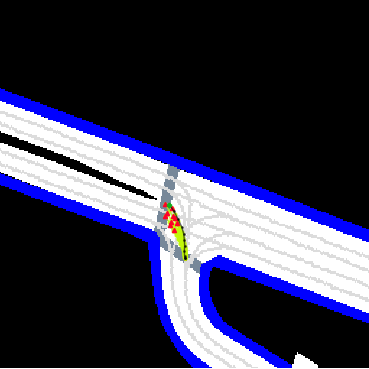}
        \caption{MTP\cite{mtp}}
    \end{subfigure}
    \begin{subfigure}{0.24\linewidth}
        \includegraphics[width=\linewidth, trim={{0.5\linewidth} {0.5\linewidth} {0.5\linewidth} {0.5\linewidth}}, clip]{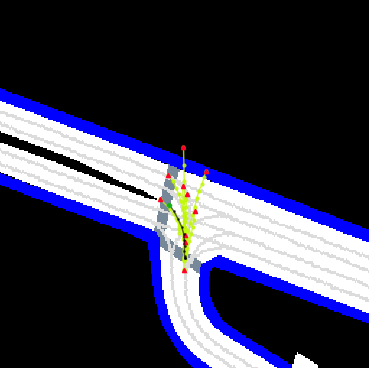}
        \caption{CoverNet\cite{covernet} Epsilon=8}
    \end{subfigure}
    \begin{subfigure}{0.24\linewidth}
        \includegraphics[width=\linewidth, trim={{0.5\linewidth} {0.5\linewidth} {0.5\linewidth} {0.5\linewidth}}, clip]{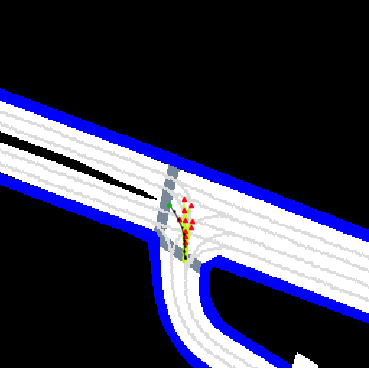}
        \caption{CoverNet\cite{covernet} Epsilon=2}
    \end{subfigure}
    \begin{subfigure}{0.24\linewidth}
        \includegraphics[width=\linewidth, trim={{0.7\linewidth} {0.7\linewidth} {0.7\linewidth} {0.7\linewidth}}, clip]{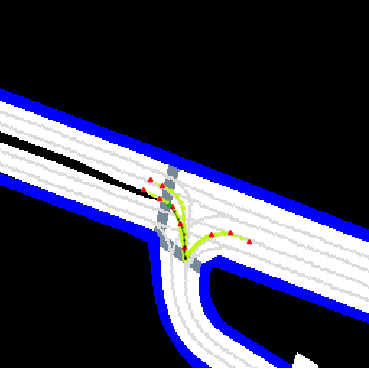}
        \caption{ALAN (Ours)}
    \end{subfigure}

    \caption{\small Shows comparison of ALAN predictions with baselines. The past trajectory is in brown and the GT is shown in black. The endpoint of GT is shown as a green dot. The predicted trajectories are shown in green and their endpoints as triangles. The final trajectories are chosen based on the predicted IOC score of each trajectory. As observed ALAN predictions are more semantically aligned in comparison.}
    \label{fig:nu_supp_img}
\end{figure*}